\documentclass[journal]{IEEEtran}

\usepackage{cite}

\ifCLASSINFOpdf
\else
\fi
\usepackage{amsmath}
\usepackage{amssymb}
\usepackage{graphicx}
\usepackage{algorithm}
\usepackage{algorithmic}
\usepackage{tikz}
\usepackage{tikz-qtree}
\usepackage{multirow}
\usepackage{rotating}
\usepackage{subfigure}
\usepackage{framed}
\usepackage{color}
\usepackage{colortbl}
\usepackage{longtable}
\usepackage{supertabular}
\definecolor{shadecolor}{rgb}{1,0,0}

\begin{document}
%
\title{A Multi- or Many-Objective Evolutionary Algorithm with Global Loop Update}
%
%
%

\author{Yingyu~Zhang~\IEEEmembership{}
\thanks{This work was supported by the National Natural Science Foundation of China under Grant 61773192.}
\thanks{Y. Zhang is with the School of Computer Science, Liaocheng University, Liaocheng 252000, China
(e-mail:zhangyingyu@lcu-cs.com).}
Bing~Zeng~\IEEEmembership{}
\thanks{B. Zeng is with the School of Software  Engineering, South China University of Technology, Guangzhou  510006, China.}
Yuanzhen~Li~\IEEEmembership{}
\thanks{Y. Li is with the School of Computer Science, Liaocheng University, Liaocheng 252000, China.}
Junqing~Li~\IEEEmembership{}
\thanks{J. Li is with the School of information science and engineering, Shandong Normal University, Jinan  250014, China, and also with the School of Computer Science, Liaocheng University, Liaocheng 252000, China.}
}

\maketitle

\begin{abstract}
Multi- or many-objective evolutionary algorithms(MOEAs), especially the decomposition-based MOEAs have been widely concerned in recent years.
The decomposition-based MOEAs emphasize convergence and diversity in a simple model and have made a great success in dealing with theoretical and practical multi- or many-objective optimization problems.
In this paper, we focus on update strategies of the decomposition-based MOEAs, and their criteria for comparing solutions.
Three disadvantages of the decomposition-based MOEAs  with local update strategies and several existing criteria for comparing solutions are analyzed and discussed.
And a global loop update strategy and two hybrid criteria are suggested.
Subsequently, an evolutionary algorithm with the global loop update is  implemented
and compared to several of the best multi- or many-objective optimization algorithms on two famous
unconstraint test suites with up to 15 objectives.
Experimental results demonstrate that unlike evolutionary algorithms with local update strategies,
the population of our algorithm does not degenerate at any generation of its evolution,
which guarantees the diversity of the resulting population.
In addition, our algorithm wins in most instances of the two test suites,
indicating that it is very competitive in terms of convergence and diversity.
Running results of our algorithm with different criteria for comparing solutions are also compared.
Their differences are very significant, indicating that the performance of our algorithm is affected by the criterion it adopts.

\end{abstract}


%
\IEEEpeerreviewmaketitle

\section{Introduction}
%
%
%
%
\IEEEPARstart{A}{} lot of  real-world problems such as electric power system reconfiguration problems\cite{Panda2009937},
water distribution system design or rehabilitation problems\cite{WBS2013}, automotive engine calibration problems\cite{AEC2013}, land use management problems\cite{LUM2012}, optimal design problems \cite{Ganesan2015293,Ganesan2013,DomingoPerez201695},
and problems of balancing between performance and cost in energy systems\cite{Najafi201446}, etc.,
can be formulated into multi- or many-objective optimization problems(MOPs) involving more than one objective function. MOPs have attracted extensive attention in recent years and
different kinds of algorithms for solving them have been proposed.
Although  algorithms based on particle swarm optimization\cite{PSO} and simulated annealing\cite{Suman2006}
developed to solve MOPs are not ignorable,
multi- or many-objective evolutionary algorithms(MOEAs) are more popular and representative in solving MOPs, such as the non-dominated sorting genetic algorithm-II (NSGA-II)\cite{NSGAII},
the strength pareto evolutionary algorithm 2(SPEA-2)\cite{SPEA2},
and the multi-objective evolutionary algorithm based on decomposition(MOEA/D)\cite{MOEAD},etc.
In General, MOEAs can be divided into three categories\cite{Survey2016}.  The first category is known as the indicator-based MOEAs.
In an indication-based MOEA, the fitness of an individual is usually evaluated by a performance indicator
such as hypervolume\cite{Emmerich2005}.
Such a performance indicator is designed to measure the convergence and diversity of the MOEA,
and hence expected to drive the population of the MOEA to converge to the Pareto Front(PF) quickly with good distribution.
The second category is  the domination-based MOEAs, in which the domination principle plays a key role.
However, in the domination-based MOEAs, other measures  have to be adopted  to maintain  the population diversity.
In NSGA-II, crowding distances of all the individuals are calculated at each generation and used to keep the population diversity ,
while reference points are used in NSGA-III\cite{NSGAIII}.
The third category  is the decomposition-based MOEAs.
In a decomposition based MOEA, a MOP is decomposed  into a set of subproblems and then optimized simultaneously.
A uniformly generated set of weight vectors associated with a fitness assignment method such as the weighted sum approach,
the Tchebycheff approach and the penalty-based boundary intersection(PBI) approach,  is usually used to decompose a given MOP.
Generally, a weight vector determines a subproblem  and defines a neighborhood.
Subproblems in a neighborhood are expected to own similar solutions and might be updated by a newly generated solution.
The decomposition-based MOEA framework emphasizes the convergence and diversity of the population in a simple model.
Therefore, it was studied extensively and improved from different points of view \cite{Carvalho2012,Ray2013,Tam2016,RVEA,MOEADD,Chen2017,Survey2017}
since it was first proposed by Zhang and Li in 2007\cite{MOEAD}.

Recently, some efforts have been made to blend  different ideas appeared in the domination-based MOEAs
and the decomposition-based MOEAs. For examples, an evolutionary many-objective optimization algorithm based on dominance
and decomposition(MOEA/DD) is proposed in \cite{MOEADD}, and a reference vector guided evolutionary algorithm is proposed in \cite{RVEA}.
In MOEA/DD, each individual is associated with a subregion uniquely determined by a weight vector,
and each weight vector (or subregion) is assigned to a neighborhood.
In an iterative step, mating parents is chosen from the neighboring subregions of the current weight vector with a given probability $\delta$,
or the whole population with a low probability $1-\delta$. In case that no associated individual exists in the selected subregions,
mating parents are randomly chosen from the whole population.
And then serval classical genetic operators such as the simulated binary crossover(SBX)\cite{SBX} and the polynomial mutation\cite{PM},etc.,
are applied on the chosen parents to generate an offspring.
Subsequently, the offspring is used to update the current population
according to a complicated but well-designed rule based on decomposition and dominance.

In this paper, we focus on update strategies of the decomposition-based evolutionary algorithms
and the criteria for comparing solutions.
Three disadvantages of the decomposition-based MOEAs with local update strategies and several existing criteria for comparing solutions are analyzed and discussed.
And a global loop update (GLU) strategy and two hybrid criteria are suggested.
Also,  we propose an evolutionary algorithm  with the GLU strategy for solving  multi- or many-objective optimization problems(MOEA/GLU).
The GLU strategy is designed to try to avoid the shortcomings of the decomposition-based MOEAs with local update strategies and eliminate bad solutions in the initial stage of the evolution,
which is expected to force the population to converge faster to the PF.

The rest of the paper is organized as follows. In section II, we provide some preliminaries used in MOEA/GLU and review serval existing criteria for comparing solutions,
i.e., PBI criterion, dominance criterion and distance criterion.
And then two hybrid criteria for judging the quality of two given solutions are suggested.
The disadvantages of the decomposition-based MOEAs with local update strategies are also analyzed in this section.
In section III, the algorithm MOEA/GLU is proposed. A general framework of it is first presented.
Subsequently, the initialization procedure,  the reproduction procedure, and the GLU procedure are elaborated.
Some discussions about advantages and disadvantages of the algorithm  are also made.
In section IV, empirical results of MOEA/GLU on DTLZ1 to DTLZ4 and WFG1 to WFG9 are compared to those of several other MOEAs, i.e., NSGA-III, MOEA/D, MOEA/DD and GrEA.
Running results of MOEA/GLU with different criteria are also compared in this section.
The paper is concluded in section V.



\section{Preliminaries and Motivations}
\subsection{MOP}
Without loss of generality, a MOP can be formulated as a minimization problem as follows:
\begin{equation}\label{MOP}
\begin{split}
Minimize \quad &F(x)=(f_1(x),f_2(x),...,f_M(x))^T \\
 &Subject \quad to \quad x\in\Omega,
 \end{split}
\end{equation}
where $M\geq 2$ is the number of objective functions, x is a decision vector,  $\Omega$ is the feasible set of decision vectors,
and $F(x)$ is composed of M conflicting objective functions. $F(x)$ is usually considered as a many-objective optimization problems when M is greater than or equal to 4.

A solution $x$ of Eq.(\ref{MOP}) is said to dominate the other one $y$ ($x\preccurlyeq y$),
if and only if $f_i(x)\leq f_i(y)$ for $i\in(1,...,M)$ and $f_j(x)<f_j(y)$ for at least one index $j\in(1,...,M)$.
It is clear that x and y are non-dominated with each other, when both $x\preccurlyeq y$ and $y \preccurlyeq x $ are not satisfied.
A solution x is Pareto-optimal to Eq.(\ref{MOP}) if there is no solution $y\in\Omega$  such
that $y\preccurlyeq x$. F(x) is then called a Pareto-optimal objective vector. The
set of all the Pareto optimal objective vectors is the PF\cite{PF}.
The goal of a MOEA is to find a set of solutions, the corresponding objective vectors of which are approximate to the PF.

\subsection{Criteria for Comparing Solutions}
\subsubsection{Dominance criterion}
Dominance is usually used to judge whether or not one solution is better than the other  in the dominance-based MOEAs.
As a criterion for comparing two given solutions, dominance can be described as follows.

\textbf{\emph{Dominance criterion:A solution x is considered to be better than the other one y when $x\preccurlyeq y$.}}

As it is discussed in \cite{Hisao2008}, the selection pressure exerted by the dominance criterion is weak in a dominance-based MOEA, and becomes weaker as the number of the objective functions increases.
It indicates that such a criterion is too stringent for MOEAs to choose the better one from two given solutions.
Therefore, in practice, the dominance criterion is usually used together with other measures.

\subsubsection{PBI criterion}
In a decomposition-based MOEA, approaches used to decompose a MOP into subproblems can be considered as criteria for comparing two solutions, such as  the weighted sum approach,
the Tchebycheff approach and the PBI approach\cite{MOEAD}.
Here, we describe the PBI approach as a criterion for comparing two given solutions.

\textbf{\emph{PBI criterion:A solution x is considered to be better than the other one y when $PBI(x)<PBI(y)$
  , where $PBI(\bullet)$ is defined as
$PBI(x)=g^{PBI}(x|w,z^{*})$, $\omega$ is a given weight vector, and $z^*$ is the ideal point.}}

The PBI function can be elaborated as\cite{MOEAD}:
\begin{equation}\label{PBI}
\begin{split}
Minimize \quad &g^{PBI}(x|w,z^{*})=d_1+\theta d_2  \\
&Subject \quad to \quad x\in \Omega
\end{split}
\end{equation}
where
\begin{equation}\label{TwoDists}
\begin{split}
&d_1=\frac{\left\|(F(x)-z^{*})^{T}w\right\|}{\|w\|}\\
&d_2=\left\|F(x)-\left(z^{*}+d_1\frac{w}{\|w\|}\right)\right\|,
\end{split}
\end{equation}
and $\theta$ is a used-defined constant penalty parameter.
In a decomposition-based MOEA with the PBI criterion,
the set of the weight vectors is usually generated at the initialization stage by the systematic sampling approach and remains unchanged in the running process of the algorithm.
The ideal point is also set at the initialization stage, but can be updated by every newly generated offspring.

\subsubsection{Distance criterion}
In \cite{IDBEA}, a criterion with respect to the two Euclidean distances $d_1$ and $d_2$ defined by Eq.(\ref{TwoDists}) are used to judge whether or not a solution is better than the other.
Denote the two Euclidean distances of x and y as $\{d_{1x},d_{2x}\}$  and $\{d_{1y},d_{2y}\}$ ,respectively.
A criterion for comparing two given solutions with respect to the two distances can be written as follows.

\textbf{\emph{Distance criterion:A solution x is considered to be better than the other one y when $d_{2x}<d_{2y}$. In the case that $d_{2x}=d_{2y}$, x is considered to be better than y when $d_{1x}<d_{1y}$.}}

\subsubsection{Two Hybrid Criteria}
It has been shown that the dominance criterion can be a good criterion for choosing better solutions in conjunction with other measures\cite{NSGAII,NSGAIII} .
And likely, the PBI criterion has achieved great success in MOEAs\cite{MOEAD,MOEADD}.
However, there are two facts with respect to these two criteria respectively can not be ignored.
The first one is that using dominance comparison alone can not exert too much selection pressure on the current population,
and hence, can not drive the population to converge to the PF of a given MOP quickly.
The second one is that it is not necessarily $PBI(x)<PBI(y)$ when $x\preccurlyeq y$, and vice versa.

Therefore, it might be natural to combine these two criteria in consideration of the two facts.
Here, we suggest two hybrid criteria.

\textbf{\emph{H1 criterion:
One solution x is considered to be better than the other one y when $x\preccurlyeq y$.
In the case that the two solutions do not dominate with each other, x is considered to be better than y when $PBI(x)<PBI(y)$.
}}

\textbf{\emph{H2 criterion:
One solution x is considered to be better than the other one y when $x\preccurlyeq y$.
In the case that the two solutions do not dominate with each other,  x is considered to be better than y when $d_{2x}<d_{2y}$.
}}

It is clear that the H1 criterion combines dominance  with the PBI criterion,
while the H2 criterion associates dominance with the Euclidean distance d2.

\subsection{The Systematic Sampling Approach}
The systematic sampling approach proposed by Das and Dennis\cite{SystematicApproach}
is usually used to generate weight vectors in MOEAs.
In this approach, weight vectors are sampled from a unit simplex.
Let $\omega=(\omega_1,...,\omega_M)^T$ is a given weight vector, $\omega_j(1\leqslant j\leqslant M)$ is the $jth$ component of $\omega$,
$\delta_j$ is the uniform spacing between two consecutive $\omega_j$ values, and $1/\delta_j$ is an integer.
The possible values of $\omega_j$ are sampled from $\{0,\delta_j,...,K_j\delta_j\}$, where $K_j=(1-\sum_{i=1}^{j-1}\omega_i)/\delta_j$.
In a special case, all $\delta_j$ are equal to $\delta$.
To generate a weight vector, the systematic sampling approach starts with sampling from $\{0,\delta,2\delta,...,1\}$ to obtain the first component $\omega_1$, and
then from $\{0,\delta,2\delta,...,K_2\delta\}$ to get the second component $\omega_2$  and so forth, until the $Mth$ component $\omega_M$
is generated. Repeat such a process, until a total of
\begin{equation}\label{nWeightVectors}
N(D,M)=\left(
\begin{array}{c}
  D+M-1 \\
  M-1
\end{array}
\right)
\end{equation}
different weight vectors  are generated, where $D > 0$ is the number of divisions considered along each objective coordinate.

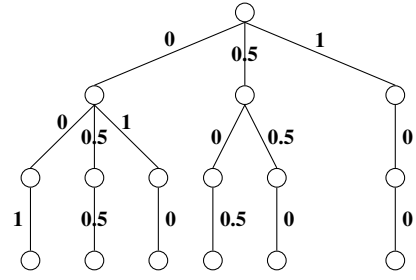
\begin{figure}
\footnotesize
\begin{center}
\begin{tikzpicture}
[level distance=11mm,
every node/.style={inner sep=2.5pt},
level 1/.style={sibling distance=20mm},
level 2/.style={sibling distance=8.5mm},
level 3/.style={sibling distance=4mm}]
\node[circle,draw] {}
child {node[circle,draw] {}
        child {node[circle,draw] {}
                    child {node[circle,draw] {}
                            edge from parent
                            node[left]{\textbf{1}}
                            }
            edge from parent
            node[left,above]{\textbf{0}}
            }
        child {node[circle,draw] {}
                    child {node[circle,draw] {}
                        edge from parent
                        node[]{\textbf{0.5}}
                        }
            edge from parent
            node[]{\textbf{0.5}}
            }
        child{node[circle,draw] {}
               child{node[circle,draw] {}
                    edge from parent
                    node[right]{\textbf{0}}
                    }
            edge from parent
            node[right,above]{\textbf{1}}
            }
        edge from parent
        node[left,above]{\textbf{0}}
    }
child {node[circle,draw] {}
            child {node[circle,draw] {}
                        child {node[circle,draw] {}
                            edge from parent
                            node[right]{\textbf{0.5}}
                            }
                    edge from parent
                    node[left]{\textbf{0}}
                    }
            child {node[circle,draw] {}
                        child {node[circle,draw] {}
                            edge from parent
                            node[right]{\textbf{0}}
                            }
                edge from parent
                node[right]{\textbf{0.5}}
                }
        edge from parent
        node[]{\textbf{0.5}}
        }
child{node[circle,draw] {}
        child{node[circle,draw] {}
                child {node[circle,draw] {}
                        edge from parent
                        node[right]{\textbf{0}}
                    }
                edge from parent
                node[right]{\textbf{0}}
            }
    edge from parent
    node[right,above]{\textbf{1}}
    };
\end{tikzpicture}
\end{center}
\caption{Generating weight vectors for $\delta=0.5$ and $M=3$ using the systematic sampling approach.}
\label{SSApproach}
\end{figure}

The approach can be illustrated by Fig.\ref{SSApproach},
in which each level represents one component of $\omega$,
and each path from the root to one of the leaves represents a possible weight vector.
Therefore, all weight vectors included in the tree can be listed as follows.
\begin{equation}
\begin{array}{lcr}
  (0  ,& 0  ,& 1  ) \\
  (0  ,& 0.5,& 0.5) \\
  (0  ,& 1  ,& 0  ) \\
  (0.5,& 0  ,& 0.5) \\
  (0.5,& 0.5,& 0  ) \\
  (1  ,& 0  ,& 0  )
\end{array}
\end{equation}

A recursive algorithm for MOEAs to generate weight vectors using the systematic sampling approach can be found in section III.
Here, we consider two cases of D taking a large value and a small value respectively.
As discussed in \cite{MOEADD} and \cite{SystematicApproach}, a large D would add more computational burden to a MOEA,
and a small D would be harmful  to the population diversity.
To avoid this dilemma, \cite{NSGAIII} and \cite{MOEADD} present a two-layer weight vector generation method.
At first, a set of $N_1$ weight vectors in the boundary layer and a set of $N_2$ weight vectors in the  inside layer are generated,
according to the systematic sampling approach described above.
Then, the coordinates of weight vectors in the inside layer are shrunk by a coordinate transformation as
\begin{equation}
v^{j}_{i}=\frac{1-\tau}{M}+\tau\times \omega^{j}_{i},
\end{equation}
where $\omega^{j}_{i}$ is the ith component of the jth weight vectors in the  inside layer, and $\tau\in [0,1]$ is a shrinkage factor set as
$\tau=0.5$ in \cite{NSGAIII} and \cite{MOEADD}.
At last, the two sets of weight vectors are combined to form the final set of weight vectors.
Denote the numbers of the weight vectors generated in the boundary layer and the inside layer as D1 and D2 respectively.
Then, the number of the weight vectors generated by the two-layer weight vector generation method is
$N(D1,M)+N(D2,M)$.

\subsection{Local update and its advantages}
Most of the decomposition-based MOEAs update the population with an offspring generated
by the reproduction operators to replace the individuals worse than the offspring in the current neighborhood.
Such an update strategy can be named as a local update(LU) strategy since it involves only the individuals in the current neighborhood.
The decomposition-based MOEAs with the LU strategy have at least two advantages.
The first one is that the LU strategy can help the algorithms to converge to the PF faster than other algorithms with non-local update strategies,
which helps them achieve great success on a lot of MOPs in the past ten years.
The second one is that the time complexities of the decomposition-based MOEAs are usually lower than those of MOEAS with non-local update strategies.
This allows them to have a great advantage in solving complicated problems or MOPs with many objectives,
since  the running time taken by a MOEA to solve a given MOP  becomes much longer as the number of the objective functions increases.

\begin{figure}[!htbp]
\begin{center}                                                       
\subfigure[]{                    
\begin{minipage}{7cm}
\centering                                                          
\includegraphics[scale=0.5]{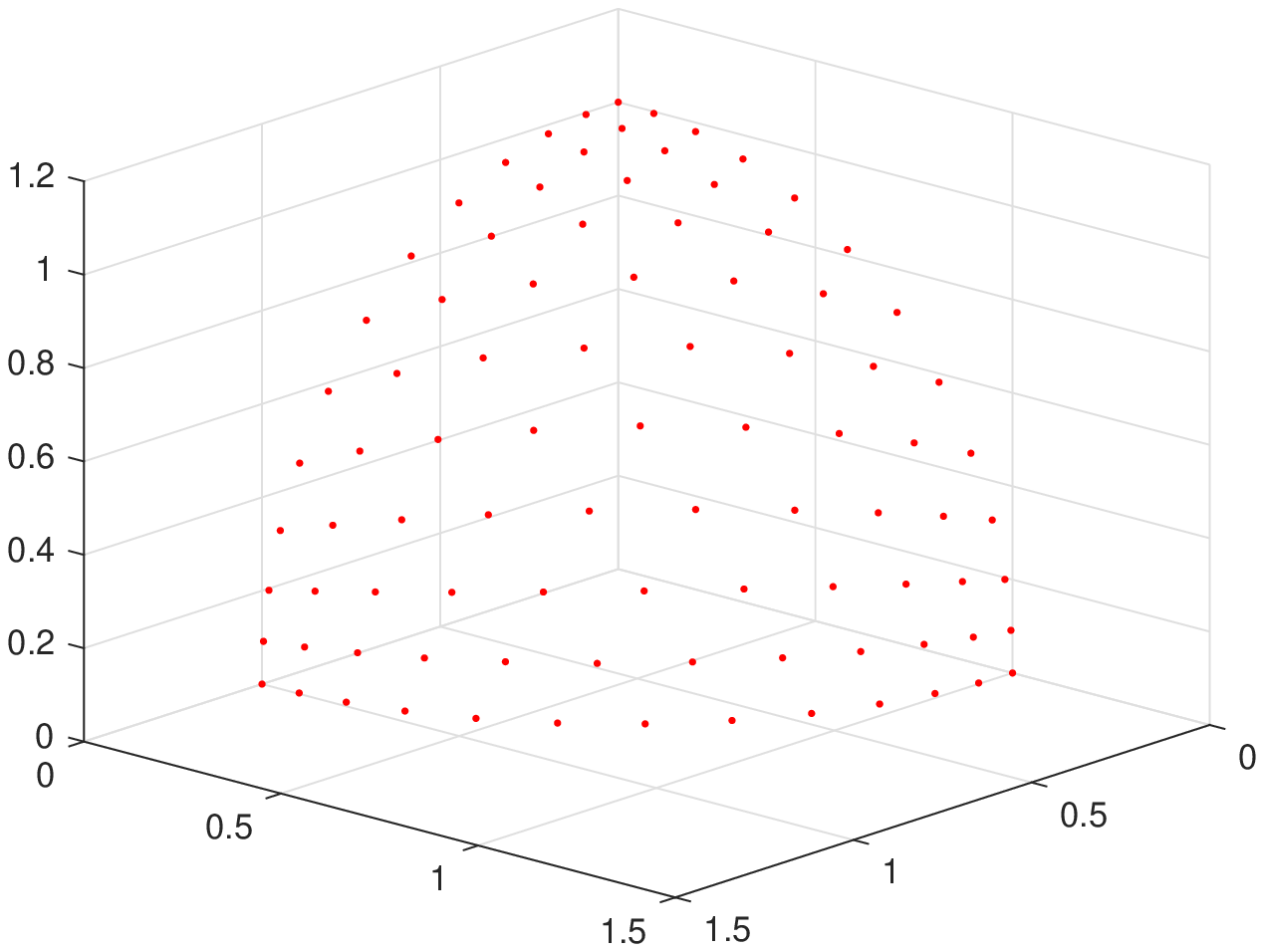}               
\end{minipage}
}

\subfigure[]{                    
\begin{minipage}{7cm}
\centering                                                          
\includegraphics[scale=0.5]{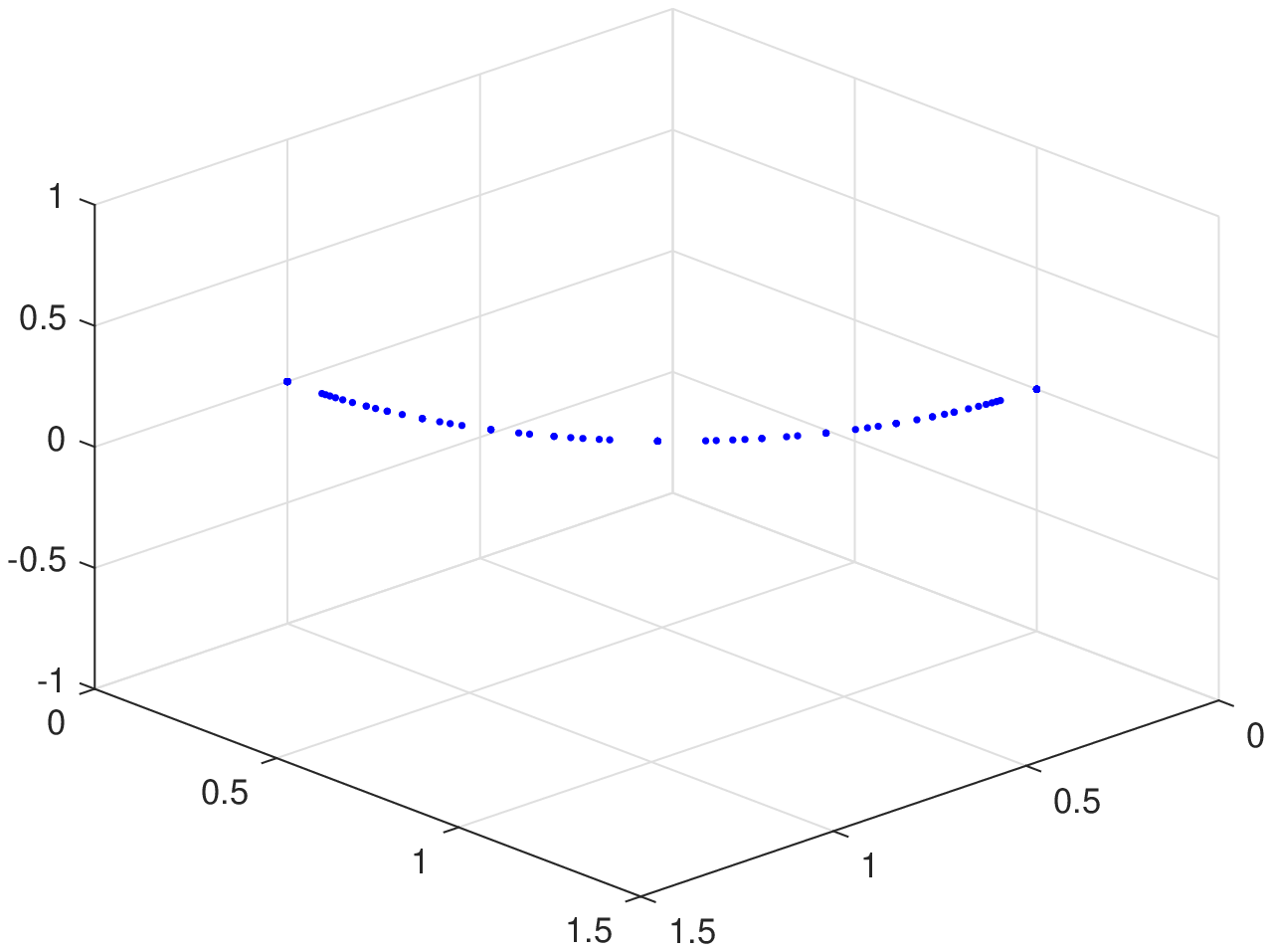}                
\end{minipage}
}
\end{center}

\caption{Two running instances of MOEA/D on DTLZ4. The first one obtained a well distributed set of solutions, and the second one obtained a degenerate set of solutions locating on an arc of a unit circle.}                       
\label{MOEADonDTLZ4}                                                        
\end{figure}

In spite of the above advantages, the decomposition-based MOEAs with the LU strategy have their own disadvantages.
The first disadvantage is that when the algorithms deal with some problems such as DTLZ4, the population may lose its diversity.
As we can see from Fig.\ref{MOEADonDTLZ4}, a running instance of MOEA/D  on DTLZ4 with 3 objectives generates well-distributed results,
while the solution set of the other one degenerates nearly to an arc on a unit circle.
What's worse, the solution set of some running instances of MOEA/D even degenerates to a few points on a unit circle in our experiments.

Notice that a call of the LU procedure replaces all individuals worse than the newly generated offspring within the current neighborhood,
which might be the reason resulting in the loss of the population diversity.
Therefore, to avoid the loss of the population diversity, one can modify the LU procedure to replace at most one individual at a call.
But the problem is how to decide which one individual is to be replaced
when there are multiple individuals  worse than the newly generated offspring.
One of the simplest replacement policies is to randomly choose one individual in the current neighborhood
and judge whether or not the offspring is better than it.
If the selected individual is worse, it will be replaced by the offspring. Or else, the offspring will be abandoned.

\begin{figure}[htbp]
\begin{center}                                                          
\subfigure[]{                    
\begin{minipage}{7cm}
\centering                                                          
\includegraphics[scale=0.5]{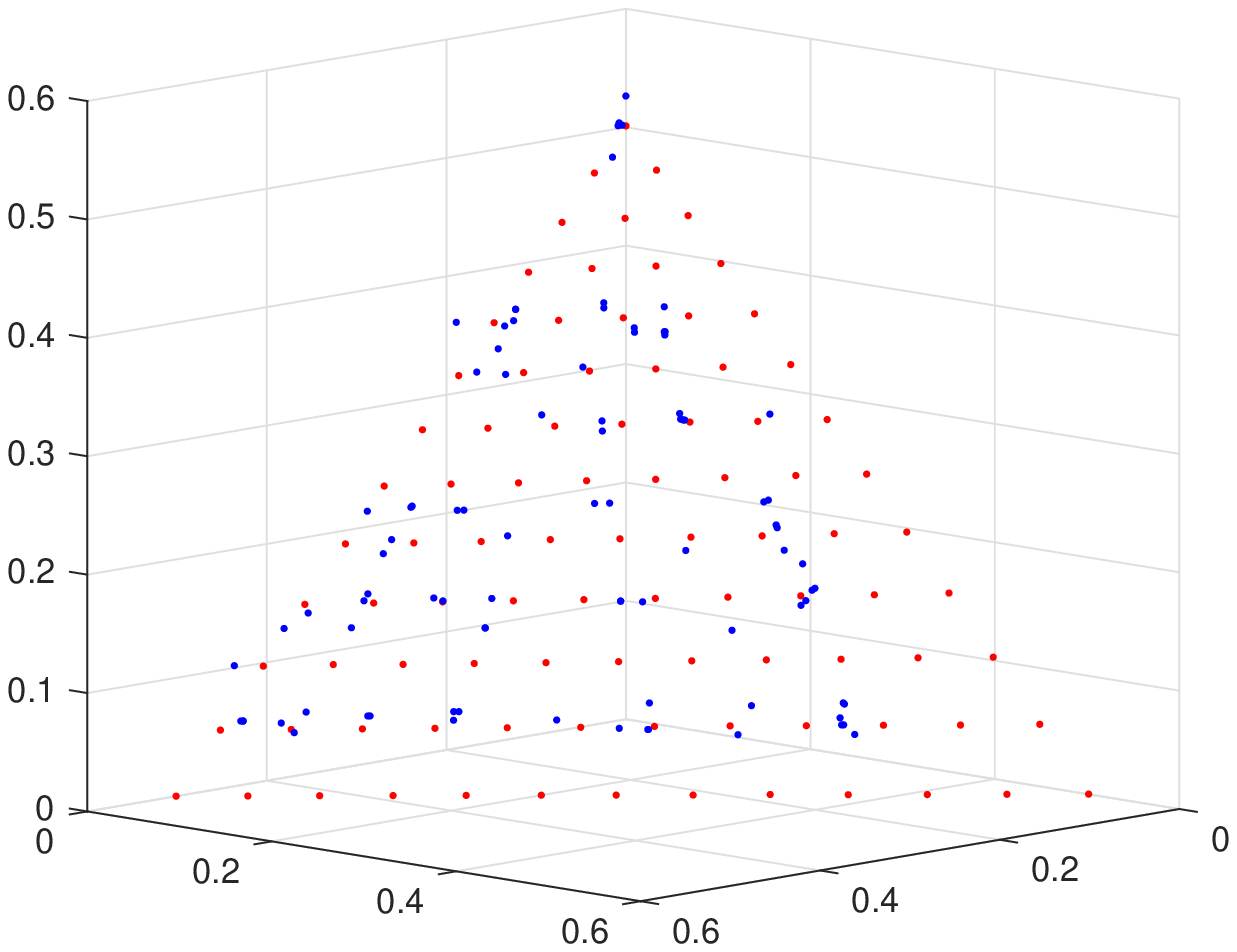}               
\end{minipage}
}

\subfigure[]{                    
\begin{minipage}{7cm}
\centering                                                          
\includegraphics[scale=0.5]{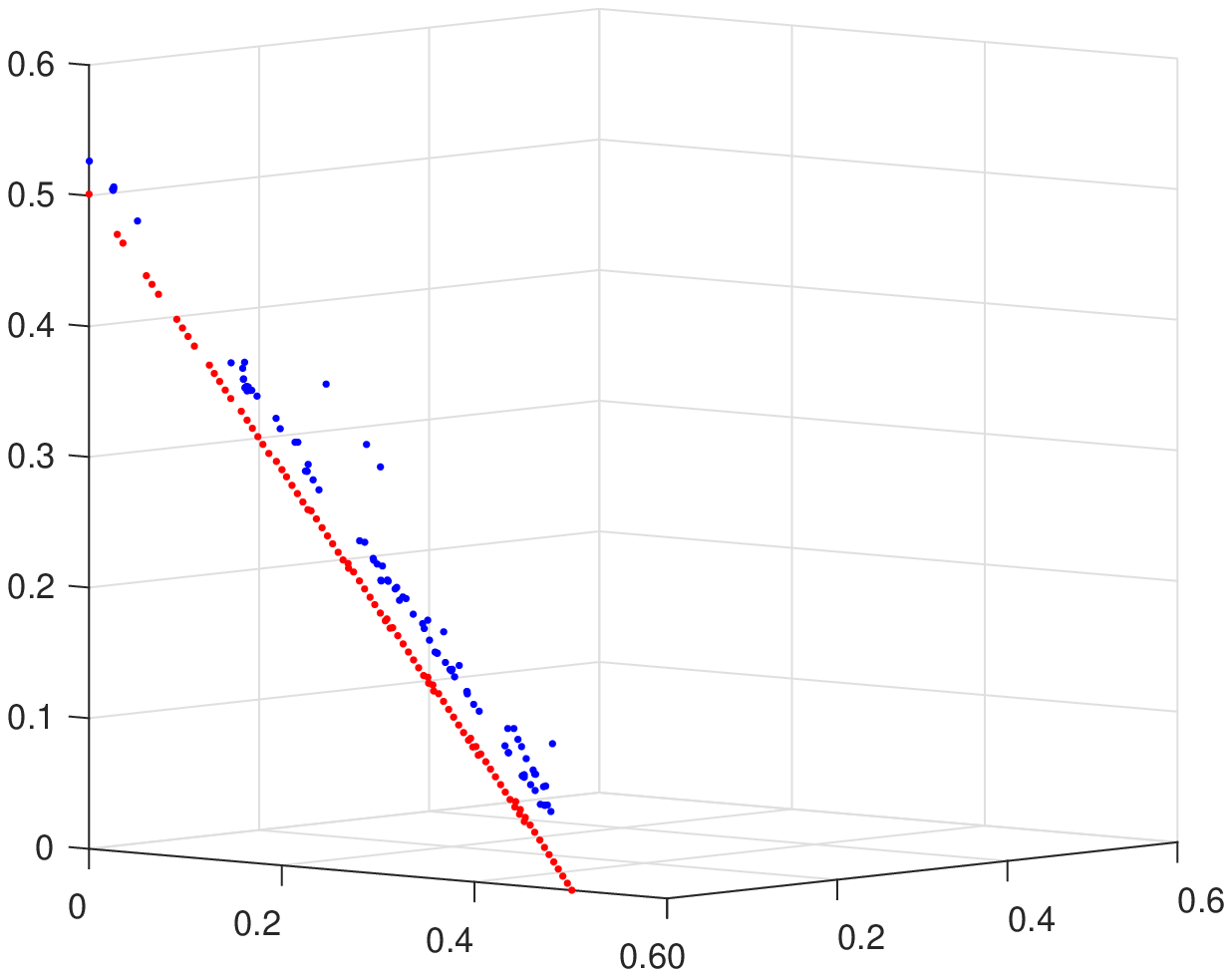}                
\end{minipage}
}
\end{center}

\caption{Running results of the original MOEA/D and its modified version with a modified LU on the DTLZ1 problem with 3 objectives. The red dots are the resulting solutions of the original MOEA/D, and the blue ones are the resulting solutions of its modified version.
Rotate (a) around the Z axis by about 90 degrees to get (b).}                       
\label{MOEADAndItsMV}                                                        
\end{figure}

Fig.\ref{MOEADAndItsMV} shows the results of the original MOEA/D and its modified version with the modified LU strategy described above on  DTLZ1  with 3 objectives.
As it can been seen from Fig.\ref{MOEADAndItsMV},
the modified LU strategy lowers down the convergence speed of the algorithm,  indicating that it is not a good update strategy.

\begin{figure}[htbp]
\begin{center}
\begin{tikzpicture}[scale=4]
\draw [->](0,0)--(0,1);
\draw [->](0,0)--(1,0);
\draw [dotted] [->](0,0)--(0.966,0.259);
\draw [dotted] [->](0,0)--(0.866,0.5);
\draw [dotted] [->](0,0)--(0.707,0.707);
\draw [dotted] [->](0,0)--(0.5,0.866);
\draw [dotted] [->](0,0)--(0.259,0.966);
\draw  [rotate=-45,red,thick](0,0.65) ellipse [x radius=0.3, y radius=0.08];
\draw [thick][red]  [<-](0.6,0.45)--(0.85,0.6);
\node [above] at (0.85,0.6) {\textcolor[rgb]{1.00,0.00,0.00}{\emph{current neighborhood}}};
\draw  [fill] (0.65,0.05) circle [radius=0.015];
\draw  [fill] (0.65,0.2) circle [radius=0.015];
\draw  [fill] (0.6,0.3) circle [radius=0.015];
\draw  [fill] (0.4,0.45) circle [radius=0.015];
\draw  [fill] (0.35,0.6) circle [radius=0.015];
\draw  [fill] (0.2,0.55) circle [radius=0.015];
\node [above] at (0.2,0.55){\textbf{A}} ;
\draw  [fill] (0.05,0.65) circle [radius=0.015];

\draw  [fill,red] (0.1,0.4) circle [radius=0.015];
\end{tikzpicture}
\end{center}
\caption{Illustration of the second disadvantage of the decomposition-based MOEAs with the LU strategy. The black dots represents the individuals in the current population, the red dot is a newly generated individual.}                                                         
\label{secondDisadvantage}                                                        
\end{figure}
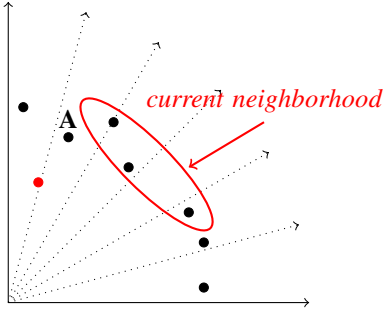

The second disadvantage of the decomposition-based MOEAs with the LU strategy is that they don't consider the individuals beyond the current neighborhood.
As we can see, such a LU strategy allows the MOEAS to update the population with less time, but it might ignore some important information leading to better convergence.
Fig.\ref{secondDisadvantage} illustrates this viewpoint.
Although the newly generated individual is better than individual A,
it will be abandoned by the decomposition-based MOEAs with the LU strategy,
since it is only compared to the individuals in the current neighborhood.

Asafuddoula et al have noticed this disadvantage of the decomposition-based MOEAs with the LU strategy\cite{IDBEA}.
The update strategy of their algorithm involves all of the individuals in the current population,
which has been demonstrated to be effective on the DTLZ and WFG test suites to some extent.
We call such an update strategy a global update(GU) strategy,
since each call of the update procedure considers all the individuals in the population, and replaces at most one individual.
In Fig.\ref{secondDisadvantage},
individual A will be replaced by the newly generated individual if a decomposition-based MOEA adopts the GU strategy instead of the LU strategy.

The third disadvantage of the decomposition-based MOEAs with the LU strategy relates to the individuals and their attached weight vectors.
As a simple example, consider the case where an individual x  and a newly generated offspring c are attached to a weight vector $\omega_x$, and an individual y is attached to $\omega_y$, so that $g^{PBI}(c|w_x,z^{*})<g^{PBI}(x|w_x,z^{*})$ and $g^{PBI}(x|w_y,z^{*})<g^{PBI}(y|w_y,z^{*})$ are satisfied.
In other words, c is better than x and x is better than y, when the weight vector $\omega_x$  and the weight vector $\omega_y$ are considered as the reference weight vector respectively.
Therefore, x will be replaced by c in a typical decomposition-based MOEA.
But so far, there is no decomposition-based MOEA  considering x as a replacement for y.

In order to deal with the three disadvantages of the decomposition-based MOEAs with the LU strategy,
we propose a MOEA with the  GLU strategy(i.e. MOEA/GLU) mentioned before, which is presented in Section III.

\section{Proposed Algorithm-MOEA/GLU}
\subsection{Algorithm Framework}
The general  framework  of MOEA/GLU is presented in Algorithm \ref{algFramework}.
As it is shown in the general framework, a \emph{while} loop is executed after the initiation procedure,
in which a $for$ loop is included.
In the $for$ loop, the algorithm runs over N weight vectors ,generates an offspring for each weight vector in the reproduction procedure, and updates the population with the offspring in the GLU procedure.

\begin{algorithm}
\caption{General Framework of MOEA/GLU}
\label{algFramework}
\begin{algorithmic}[1]
\ENSURE  Final Population.
\STATE   Initialization Procedure.
\WHILE{The stop condition is not stisfied}
\FOR{$i=1$ to N}
\STATE Reproduction Procedure.
\STATE The GLU procedure
\ENDFOR
\ENDWHILE
\end{algorithmic}
\end{algorithm}

\subsection{Initialization Procedure}
The initialization procedure includes four steps.
In the first step, a set of uniformly distributed weight vectors are generated using the systematic approach proposed in \cite{SystematicApproach}.
A recursive algorithm for generating the weight vectors is presented in algorithm \ref{algSSA} and \ref{RecursiveBodyofSSA}.
\begin{algorithm}
\caption{The systematic sampling approach}
\begin{algorithmic}[1]
\label{algSSA}
\REQUIRE D:the number of divisions, M:the number of objectives.
\ENSURE  A set of uniform weight vectors.
\STATE $\omega=(0,...,0)$;
\STATE $Gen\_ith\_Level(\omega,0,0,D,M)$;
\end{algorithmic}
\end{algorithm}

Algorithm \ref{algSSA} calls the recursive function $Gen\_ith\_Level$ described in algorithm \ref{RecursiveBodyofSSA}
with $\omega=(0,...,0)$, $K=0$, and $i=0$, to generate weight vectors.
At the ith level of the recursive function,
the ith component of a weight vector is generated.
As discussed before, the value of each component of a weight vector ranges from 0 to 1 with the step size $1/D$, and all components of a weight vector sum up to 1.
In other words, all components of a weight vector share D divisions.
Therefore, if K=D(K is the number of divisions that have been allocated), then the rest of the components are all set to zero.
In addition, if $\omega[i]$ is the last component, i.e., $i=M-1$, then all the remaining divisions are assigned to it.
Both the two cases indicate the end of a recursive call, and a generated weight vector is output.

\begin{algorithm}
\caption{$Gen\_ith\_Level(\omega,K,i,D,M)$}
\begin{algorithmic}[1]
\label{RecursiveBodyofSSA}
\IF{k==D}
    \STATE $\omega[i],...,\omega[M-1]\leftarrow 0$;
    \STATE output($\omega$);
    \STATE return;
\ENDIF
\IF{i==M-1}
    \STATE $\omega[i]\leftarrow (D-K)/D$;
    \STATE output($\omega$);
    \STATE return;
\ENDIF
\FOR{j=0 to D-K}
\STATE　$\omega[i]\leftarrow j/D$;
\STATE $Gen\_ith\_Level(\omega,K+j,i+1,D,M)$;
\ENDFOR
\end{algorithmic}
\end{algorithm}

One of the main ideas of MOEA/GLU is that each individual is attached to a weight vector and a weight vector owns only one individual.
Meanwhile, each weight vector determines a neighborhood.
In the second step, the neighborhoods of all weight vectors are generated by calculating the Euclidean distances of the weight vectors using Eq.(\ref{TwoDists}).
Subsequently, a population of N individuals is initialized randomly and attached to
N weight vectors in order of generation in the third step.
Finally, the ideal point is initialized in the fourth step, which can be updated by every offspring in the course of evolution.

\subsection{Reproduction Procedure}
The reproduction procedure can be described as follows.
Firstly, a random number r between 0 and 1 is generated. If r is less than a given selection probability $P_s$,
then choose two individuals from the neighborhood of the current weight, or else choose two individuals from the whole population.
Secondly, the SBX operator is applied  on the two individuals to generate two intermediate individuals.
Notice that, if both of the two individuals are evaluated and used to update the population,
then the number of  individuals evaluated at each generation will be twice as many as that of the individuals in the whole population.
However, the number of individuals evaluated at each generation in many popular MOEAs such as NSGA-III and MOEADD etc.,
is usually the same as the size of the population. Therefore, one of the two intermediate individuals is abandoned at random for the sake of fairness.
Finally, the polynomial mutation operator is applied on the reserved intermediate individual to generate an
offspring, which will be evaluated and used to update the current population in the following GLU procedure.

\subsection{The GLU procedure}
\begin{algorithm}
\caption{The GLU procedure}
\label{GLU}
\begin{algorithmic}[1]
\REQUIRE a new offspring c, the current population P.
\STATE bFlag=true;
\WHILE{bFlag}
\STATE $Find\_Attathed\_Weight(c)\rightarrow i$;
\IF{c is better than P[i]}
\STATE Swap(c,P[i]);
\ELSE
\STATE bFlag=false;
\ENDIF
\ENDWHILE


\end{algorithmic}
\end{algorithm}

The GLU procedure is illustrated in Algorithm \ref{GLU}, which can described as follows.
Each individual is attached to a weight vector, which has the shortest perpendicular distance to the weight vector.
$Find\_Attathed\_Weight(c)$ is designed to find the attached weight of c, in which the perpendicular distance is calculated by Eq.(\ref{TwoDists}).
Denote the perpendicular distance of the $ith$ individual $P[i]$  to the $jth$  weight vector as $d_{ij}$.
A given weight vector maintains only one slot to keep the best individual attached to it generated so far from the beginning of the algorithm.
The minimum value of $\{d_{i1},d_{i2},...,d_{iN}\}$ can be expected to be $d_{ii}$ after the algorithm evolves enough generations.
However, in the initialization stage, all the individuals are generated randomly,
and attached to the weight vectors in order of generation.
In other words, the $ith$ weight vector may not be the one, to which its attached individual $P[i]$ has the shortest perpendicular distance.
Supposed that the minimum value of $\{d_{i1},d_{i2},...,d_{iN}\}$ is still $d_{ij}$ at a certain generation,
and  the offspring c is better than $P[i]$ . Then, $P[i]$ will be replaced out by c, and  considered as a candidate to take the place of  the individual hold by the $jth$ weight vector, i.e., $P[j]$.


\subsection{Discussion}
This section gives a simple discussion about the similarities and differences of MOEA/GLU, MOEA/D, and MOEA/DD.
\begin{enumerate}
  \item Similarities of  MOEA/GLU, MOEA/D, and MOEA/DD.
        MOEA/GLU and MOEA/DD can be seen as two variants of MOEA/D to some extent, since all of the three algorithms employ decomposition technique to deal with MOPs.
        In addition, a set of weight vectors is used to guide the selection procedure, and the concept of neighborhood plays an important role in all of them.
  \item Differences between MOEA/GLU and MOEA/D.
        Firstly, MOEA/D uses a LU strategy, and  MOEA/GLU employs the  so-called GLU strategy, which considers all of the individuals in the current population at each call of the update procedure.
        Secondly, to judge whether or not an individual is better than the other, MOEA/D compares the fitness values of them, while other criteria for comparing individuals can also be used in MOEA/GLU.
        Thirdly, once a individual is generated in MOEA/D, all the individuals in the current neighborhood that worse than it will be replaced.
        However, each individual is attached to one weight vector in MOEA/GLU, and a newly generated individual is only compared to the old one attached to the same weight vector.
        The replacement operation occurs only when the new individual is better than the old one.
  \item Differences between MOEA/GLU and MOEA/DD.
        In the first place, one weight vector in MOEA/DD not only defines a subproblem,
        but also specifies a subregion that can be used to estimate the local density of a population.
        In principle, a subregion owns zero, one, or more individuals at any generation.
        In MOEA/GLU, each individual is attached only to one weight vector, and a weight vector can hold only one individual.
        In the second place, the dominance criterion can be taken into account in MOEA/GLU, the way that it is used is different from that of MOEA/DD.
        In MOEA/GLU, the dominance between the newly generated individual and the old one attached to the same weight vector can be used to judge which of the two is better, while the dominance criterion is considered among all individuals within a subregion  in MOEA/DD.
\end{enumerate}

\subsection{Time Complexity}
The function \emph{Find\_Attathed\_Weight} in the  GLU  procedure runs over all weight vectors,
calculates the perpendicular distances between the newly generated offspring and all weight vectors,
and finds the weight vector, to which the offspring has the shortest perpendicular distance.
Therefore, it takes $O(MN)$ times of floating-point calculations for the function \emph{Find\_Attathed\_Weight}  to find the attached weight vector of the offspring,
where M is the number of the objective functions and N is the size of the population.

As it is indicated before,
the \emph{while} loop is designed to help the individuals in the  initial stage of the algorithm to find
their attached weight vectors quickly.
The fact that the individuals at a certain generation do not attach to their corresponding weight vectors causes extra  entries into the function \emph{Find\_Attathed\_Weight}.
However, once all of the individuals are attached to their corresponding weight vectors, the function \emph{Find\_Attathed\_Weight} will be entered at most two times.
Let the entries into the function \emph{Find\_Attathed\_Weight} be $(1+N_i)$ times at each call of the GLU procedure,
and denote the number of the generations as G.
Since  $\sum_{i}{N_i}\leq N$ and  the GLU procedure is called $NG$ times in the whole process of MOEA/GLU,
the time complexity of the algorithm is $O(MN^2G)$,
which is the same as that of MOEA/DD, but worse than that of MOEA/D.

\section{Experimental Results}

\subsection{Performance Metrics}
\subsubsection{Inverted Generational Distance(IGD)}
Let S be a result solution set of a MOEA on a given MOP.
Let $R$ be a uniformly distributed representative points of the PF.
The IGD value of S relative to R can be calculated as\cite{IGD}
\begin{equation}
IGD(S,R)=\frac{\sum_{r\in R}d(r,S)}{|R|}
\end{equation}
where d(r,S) is the minimum Euclidean distance between r and the points in S, and $|R|$ is the cardinality of R. Note that, the points in R should be well distributed and $|R|$ should be large enough to ensure that the points in R could represent the PF very well. This guarantees that the IGD value of S is able to measure the convergence and diversity of the solution set. The lower the IGD value of S, the better its quality\cite{MOEADD}.

\subsubsection{HyperVolume(HV)}
The HV value of a given solution set S is defined as\cite{HV}
\begin{equation}
HV(S)=vol\left( \bigcup_{x\in S}\left[ f_1(x),z_1 \right]\times \ldots \times\left[ f_M(x),z_M \right]\right),
\end{equation}
where $vol(\cdot)$ is the Lebesgue measure,and $z^r=(z_1,\ldots,z_M)^T$ is a given reference point. As it can be seen that the HV value of S is a measure of the size of the objective space dominated by the solutions in S and bounded by $z^r$.

As with \cite{MOEADD}, an algorithm based on Monte Carlo sampling proposed in \cite{HYPE}  is applied to compute the approximate HV values for 15-objective test instances, and the WFG algorithm \cite{WFGalgorithm} is adopted to compute the exact HV values for other test instances for the convenience of comparison. In addition, all the HV values are normalized to $[0,1]$ by dividing $\prod_{i=1}^{M}z_i$.

\subsection{Benchmark Problems}
\subsubsection{DTLZ test suite}
Problems DTLZ1 to DTLZ4 from the DTLZ test suite proposed by Deb et al\cite{DTLZ}  are chosen for our experimental studies in the first place.
One can refer to \cite{DTLZ} to find their definitions.
Here, we only summarize some of their features.
\begin{itemize}
  \item DTLZ1:The global PF of DTLZ1 is the linear hyper-plane $\sum_{i=1}^{M}f_i=0.5$. And the search space contains $(11^k-1)$
  local PFs that can hinder a MOEA to converge to the hyper-plane.
  \item DTLZ2:The global PF of DTLZ2 satisfys $\sum_{i}^{M}f_i^2=1$.
  Previous studies have shown that this problem is easier to be solved by existing MOEAs,
   such as NSGA-III, MOEADD, etc., than DTLZ1, DTLZ3 and DTLZ4.
  \item DTLZ3:The definition of the glocal PF of DTLZ3 is the same as that of DTLZ2.
  It introduces $(3^k-1)$ local PFs. All local PFs are parallel to the global PF
  and a MOEA can get stuck at any of these local PFs before converging to the global PF.
  It can be used to investigate a MOEA's ability to converge to the global PF.
  \item DTLZ4:The definition of the global PF of DTLZ4 is also the same as that of DTLZ2 and DTLZ3.
  This problem can be obtained by modifying DTLZ2 with a different meta-variable mapping,
  which is expected to introduce a biased density of solutions in the search space.
  Therefore, it can be used to investigate a MOEA's ability to maintain a good distribution of solutions.
\end{itemize}

To calculate the IGD value of a result set S of a MOEA running on a MOP, a set $R$ of representative points of the PF needs to be given in advance.
For DTLZ1 to DTLZ4,  we take the set of the intersecting points of weight vectors
and the PF surface as $R$.
Let $f^*=(f_{1}^*,...,f_{M}^*) $ be the intersecting point of a weight vector $w=(w_1,...,w_M)^T$ and the PF surface.
Then $f_i^*$ can be computed as\cite{MOEADD}
\begin{equation}
f_i^*=0.5\times\frac{w_i}{\sum_{j=1}^{M}w_j}
\end{equation}
for DTLZ1, and
\begin{equation}
f_i^*=\frac{w_i}{\sqrt{\sum_{j=1}^{M}w_j}}
\end{equation}
for DTLZ2, DTLZ3 and DTLZ4.

\subsubsection{WFG test suite\cite{WFGProblems,WFG}}
This test suite allows test problem designers to construct scalable test problems with any number of objectives,
in which features such as modality and separability can be customized as required.
As discussed in \cite{WFGProblems,WFG},
it exceeds the functionality of the DTLZ test suite.
In particular, one can construct non-separable problems, deceptive problems,
truly degenerative problems, mixed shape PF problems,
problems scalable in the number of position-related parameters,
and problems with dependencies between position- and distance-related parameters as well with
the WFG test suite.

In \cite{WFG}, several scalable problems, i.e., WFG1 to WFG9,
are suggested for MOEA designers to test their algoritms,
which can be described as follows.
\begin{equation}
\begin{split}
Minimize \quad F(X)&=(f_1(X),...,f_M(X))\\
f_i(X)&=x_M+2ih_i(x_1,...,x_{M-1}) \\
X&=(x_1,...,x_M)^T
\end{split}
\end{equation}
where $h_i$ is a problem-dependent shape function  determining the geometry of the fitness
space, and $X$ is derived from a vector of working parameters $Z=(z_1,...,z_n)^T, z_i\in [0,2i]$ , by employing four problem-dependent transformation functions $t_1$, $t_2$, $t_3$  and  $t_4$.
Transformation functions must be designed carefully such that the underlying PF remains intact with a relatively easy to determine Pareto optimal set.
The WFG Toolkit provides a series of predefined shape and transformation functions to help ensure this is the case.
One can refer to  \cite{WFGProblems,WFG} to see their definitions.
Let
\begin{equation}
\begin{split}
Z''&=(z''_1,...,z''_m)^T
=t_4(t_3 (t_2 (t_1(Z'))))\\
Z'&=(z_1/2,...,z_n/2n)^T.
\end{split}
\end{equation}
Then $x_i=z''_i(z''_i-0.5)+0.5$ for problem WFG3, whereas $X=Z''$ for problems WFG1, WFG2 and WFG4 to WFG9.

The features of WFG1 to WFG9 can be summarized as follows.
\begin{itemize}
  \item WFG1:A separable and uni-modal problem with a biased PF and a convex and mixed geometry.
  \item WFG2:A non-separable problem with a convex and disconnected geometry, i.e., the PF of WFG2 is composed of several disconnected convex segments. And all of its objectives but $f_M$ are uni-modal.
  \item WFG3:A non-separable and uni-modal problem with a linear and degenerate PF shape, which can be seen as a connected version of WFG2.
  \item WFG4:A separable and multi-modal problem with large "hill sizes", and a concave geometry.
  \item WFG5:A separable and deceptive problem with a concave geometry.
  \item WFG6:A nonseparable and uni-modal problem with a concave geometry.
  \item WFG7:A separable and uni-modal problem with parameter dependency, and a concave geometry.
  \item WFG8:A nonseparable and uni-modal problem with parameter dependency, and a concave geometry.
  \item WFG9:A nonseparable, deceptive and uni-modal problem with parameter dependency, and a concave geometry.
\end{itemize}

As it can be seen from above, WFG1 and WFG7 are both separable and uni-modal,
and WFG8 and WFG9 have nonseparable property,
but the parameter dependency of WFG8 is much harder than that caused of WFG9.
In addition, the deceptiveness of WFG5 is more difficult than that of WFG9,
since WFG9 is only deceptive on its position parameters.
However, when it comes to the nonseparable reduction, WFG6 and WFG9 are more difficult than  WFG2 and WFG3.
Meanwhile,problems WFG4 to WFG9 share the same EF shape in the objective space,
which is a part of a hyper-ellipse with radii $r_i = 2i$, where $i\in\{1,...,M\}$.

\subsection{Parameter Settings}
The parameter settings of MOEA/GLU are listed as follows.
\begin{enumerate}
  \item Settings for Crossover Operator:The crossover probability is set as $p_c=1.0$ and the distribution index is $\eta_c=30$.
  \item Settings for Mutation Operator:The mutation probability is set as $p_m=0.6/n$, and is different from that of MOEA/DD, which is 1/n. The distribution index is set as $\eta_m=20$.
  \item Population Size:The population size of MOEA/GLU is the same as the number of the weight vectors that can be calculated by Eq.(\ref{nWeightVectors}). Since the divisions for 3- and 5-objective instances are set to 12 and 6, and the population sizes of them are 91 and 210, respectively. As for 8-, 10- and 15-objective instances, two-layer weight vector generation method is applied.  The divisions and the population sizes of them are listed in Table \ref{nDivisions}.
  \item Number of Runs:The algorithm is independently run 20 times on each test instance, which is the same as that of other algorithms for comparison.
  \item Number of Generations: All of the algorithms stopped at a predefined number of generations. The number of generations for DTLZ1 to DTLZ4 is listed in Table \ref{nGens}, and the number of generations for all the instances of WFG1 to WFG9 is 3000.
  \item Penalty Parameter in PBI: $ \theta= 5.0$.
  \item Neighborhood Size: $T = 20$.
  \item Selection Probability: The probability of selecting two mating individuals from the current neighborhood is set as $p_s = 0.9$.
  \item Settings for DTLZ1 to DTLZ4:As in papers\cite{IDBEA,MOEADD},
  the number of the objectives are set as $M \in \{3,5,8,10,15\}$ for comparative purpose.
  And the number of the decision variables is set as $n = M + r-1$, where $r = 5$ for DTLZ1, and $r = 10$ for DTLZ2, DTLZ3 and DTLZ4.
  To calculate the HV value  we set the reference point to $(1,...,1)^T$ for DTLZ1,  and $(2,...,2)^T$ DTLZ2 to DTLZ4.
  \item Settings for WFG1 to WFG9:
  The number of the decision variables is set as n = k + l,
  where   $k = 2\times(M-1)$  is the position-related variable and $l = 20$ is the distance-related variable.
  To calculate the HV values for problems WFG1 to WFG9, the reference point is set to $(3,...,2M+1)^T$.
\end{enumerate}

\begin{table}
\caption{Number of Population Size}
\begin{center}\label{nDivisions}
\begin{tabular}{|c|c|c|c|}
\hline
M&D1&D2&Population Size\\
\hline
3&12&-&91\\
5&6&-&210\\
8&3&2&156\\
10&3&2&275\\
15&2&1&135\\
\hline
\end{tabular}
\end{center}
\end{table}

\begin{table}
\caption{Number OF Generations}
\begin{center}
\begin{tabular}{|c|c|c|c|c|c|}

  \hline
  problem& $M=3$ & $M=5$ & $M=8$ & $M=10$ & $M=15$ \\
  \hline
 DTLZ1 & 400 & 600 & 750 & 1000 & 1500 \\
  DTLZ2 & 250 & 350 & 500 & 750 & 1000 \\
  DTLZ3 & 1000 & 1000 & 1000 & 1500 & 2000 \\
  DTLZ4 & 600 & 1000 & 1250 & 2000 & 3000 \\
  \hline
\end{tabular}\label{nGens}
\end{center}
\end{table}

\subsection{Performance Comparisons on DTLZ1 to DTLZ4}

\setlength{\fboxsep}{0.5pt}
\begin{table*}[!htbp]
\caption{Best, Median and Worst IGD Values by MOEA/GLU, MOEA/DD, NSGA-III, MOEA/D and GrEA
on DTLZ1, DTLZ2, DTLZ3 and DTLZ4 instances with Different Number of Objectives.
The values in red are the best, and the values in gray are the second best.}
\begin{center}\label{IGDonDTLZs}
\resizebox{\textwidth}{!}{ %
\begin{tabular}{|c|c|c|c|c|c|c|c|c|c|c|c|c|c}
 \hline
  &m & MOEA/GLU & MOEA/DD&NSGA-III&MOEA/D&GrEA &&MOEA/IUP&MOEA/DD&NSGA-III&MOEA/D&GrEA\tabularnewline
  \hline

  \multirow{15}{0.2cm}{\begin{sideways}{DTLZ1}\end{sideways}}
  && \colorbox{red}{1.073E-4}  &\colorbox{gray}{3.191E-4} &4.880E-4&4.095E-4&2.759E-2 &\multirow{15}{0.2cm}{\begin{sideways}{DTLZ3}\end{sideways}}&\colorbox{red}{1.598E-4} & \colorbox{gray}{5.690E-4} &9.751E-4&9.773E-4&6.770E-2 \\
  &3& \colorbox{red}{3.608E-4} & \colorbox{gray}{5.848E-4} &1.308E-3&1.495E-3&3.339E-2&& \colorbox{red}{1.257E-3} &\colorbox{gray}{1.892E-3} &4.007E-3&3.426E-3&7.693E-2\\
  && \colorbox{gray}{1.669E-3} & \colorbox{red}{6.573E-4} &4.880E-3&4.743E-3&1.351E-1&& \colorbox{gray}{8.138E-3} &\colorbox{red}{6.231E-3} &6.665E-3&9.113E-3&4.474E-1\\

  \cline{2-7} \cline{9-13}
  &&  \colorbox{red}{1.732E-4} & \colorbox{gray}{2.635E-4} &5.116E-4&3.179E-4&7.369E-2&& \colorbox{red}{2.965E-4} & \colorbox{gray}{6.181E-4} &3.086E-3&1.129E-3&5.331E-1\\
  &5& \colorbox{red}{2.115E-4} & \colorbox{gray}{2.916E-4} &9.799E-4&6.372E-4&3.363E-1&& \colorbox{red}{8.390E-4} & \colorbox{gray}{1.181E-3} &5.960E-3&2.213E-3&8.295E-1\\
  && \colorbox{red}{2.395E-4} & \colorbox{gray}{3.109E-4} &1.979E-3&1.635E-3&4.937E-1&& \colorbox{red}{2.543E-3} & \colorbox{gray}{4.736E-3} &1.196E-2&6.147E-3&1.124E+0\\

  \cline{2-7} \cline{9-13}
  && \colorbox{red}{1.457E-3} & \colorbox{gray}{1.809E-3} &2.044E-3&3.914E-3&1.023E-1&& \colorbox{red}{1.987E-3} &\colorbox{gray}{3.411E-3}&1.244E-2&6.459E-3&7.518E-1\\
  &8& \colorbox{red}{2.069E-3} & \colorbox{gray}{2.589E-3} &3.979E-3&6.106E-3&1.195E-1&& \colorbox{red}{4.478E-3} &\colorbox{gray}{8.079E-3}&2.375E-2&1.948E-2&1.024E+0\\
  && \colorbox{gray}{3.388E-3} & \colorbox{red}{2.996E-3} &8.721E-3&8.537E-3&3.849E-1&& \colorbox{red}{1.759E-2} &\colorbox{gray}{1.826E-2}&9.649E-2&1.123E+0&1.230E+0\\

  \cline{2-7} \cline{9-13}
  && \colorbox{red}{1.429E-3} & \colorbox{gray}{1.828E-3} &2.215E-3&3.872E-3&1.176E-1&& \colorbox{gray}{2.173E-3} & \colorbox{red}{1.689E-3} &8.849E-3&2.791E-3&8.656E-1\\
  &10& \colorbox{red}{2.030E-3} & \colorbox{gray}{2.225E-3} &3.462E-3&5.073E-3&1.586E-1&& \colorbox{gray}{2.663E-3} & \colorbox{red}{2.164E-3} &1.188E-2&4.319E-3&1.145E+0\\
  && \colorbox{gray}{3.333E-3} & \colorbox{red}{2.467E-3} &6.896E-3&6.130E-3&5.110E-1&& \colorbox{gray}{4.795E-3}& \colorbox{red}{3.226E-3} & 2.082E-2 &1.010E+0&1.265E+0\\

  \cline{2-7} \cline{9-13}
  && \colorbox{red}{2.261E-3} & \colorbox{gray}{2.867E-3} &2.649E-3&1.236E-2&8.061E-1&& \colorbox{red}{5.299E-3} & \colorbox{gray}{5.716E-3} &1.401E-2&4.360E-3&9.391E+1\\
  &15& \colorbox{red}{3.652E-3} & \colorbox{gray}{4.203E-3} &5.063E-3&1.431E-2&2.057E+0&& \colorbox{gray}{8.732E-3} & \colorbox{red}{7.461E-3} &2.145e-2&1.664E-2&1.983E+2\\
  && \colorbox{gray}{6.111E-3} & \colorbox{red}{4.669E-3} &1.123E-2&1.692E-2&6.307E+1&& \colorbox{gray}{1.912E-2} & \colorbox{red}{1.138E-2}& 4.195E-2&1.260E+0&3.236E+2\\
 \hline
 \hline
   \multirow{15}{0.2cm}{\begin{sideways}{DTLZ2}\end{sideways}}
  && \colorbox{red}{4.418E-4} & 6.666E-4&1.262E-3& \colorbox{gray}{5.432E-4} &6.884E-2&\multirow{15}{0.2cm}{\begin{sideways}{DTLZ4}\end{sideways}}& \colorbox{red}{9.111E-5} & \colorbox{gray}{1.025E-4} &2.915E-4&2.929E-1&6.869E-2 \\
  &3& \colorbox{red}{5.738E-4} & 8.073E-4&1.357E-3& \colorbox{gray}{6.406E-4} &7.179E-2&& \colorbox{red}{1.105E-4} & \colorbox{gray}{1.429E-4} &5.970E-4&4.280E-1&7.234E-2\\
  && \colorbox{red}{7.510E-4} & 1.243E-3&2.114E-3& \colorbox{gray}{8.006E-4} &7.444E-2&& \colorbox{red}{1.385E-4} & \colorbox{gray}{1.881E-4} &4.286E-1&5.234E-1&9.400E-1\\

 \cline{2-7} \cline{9-13}
  && \colorbox{red}{9.513E-4} & \colorbox{gray}{1.128E-3} &4.254E-3&1.219E-3&1.411E-1&& \colorbox{red}{7.218E-5} & \colorbox{gray}{1.097E-4} &9.849E-4&1.080E-1&1.422E-1\\
  &5& \colorbox{red}{1.075E-3} & \colorbox{gray}{1.291E-3} &4.982E-3&1.437E-3&1.474E-1&& \colorbox{red}{9.255E-5} & \colorbox{gray}{1.296E-4} &1.255E-3&5.787E-1&1.462E-1\\
  && \colorbox{red}{1.231E-3} & \colorbox{gray}{1.424E-3} &5.862E-3&1.727E-3&1.558E-1&& \colorbox{red}{1.115E-4} & \colorbox{gray}{1.532E-4} &1.721E-3&7.348E-1&1.609E-1\\

  \cline{2-7} \cline{9-13}
  && \colorbox{red}{2.553E-3} & \colorbox{gray}{2.880E-3} &1.371E-2&3.097E-3&3.453E-1&& \colorbox{red}{3.540E-4} & \colorbox{gray}{5.271E-4} &5.079E-3&5.298E-1&3.229E-1\\
  &8& \colorbox{red}{3.038E-3} & \colorbox{gray}{3.291E-3}&1.571E-2&3.763E-3&3.731E-1&& \colorbox{red}{4.532E-4} & \colorbox{gray}{6.699E-4} &7.054E-3&8.816E-1&3.314E-1\\
  && \colorbox{red}{3.375E-3} & \colorbox{gray}{4.106E-3}&1.811E-2&5.198E-3&4.126E-1&& \colorbox{red}{5.823E-4} & \colorbox{gray}{9.107E-4} &6.051E-1&9.723E-1&3.402E-1\\

  \cline{2-7} \cline{9-13}
  && \colorbox{gray}{2.917E-3} & 3.223E-3 &1.350E-2& \colorbox{red}{2.474E-3} &4.107E-1&& \colorbox{red}{8.397E-4}& \colorbox{gray}{1.291E-3} &5.694E-3&3.966E-1&4.191E-1\\
  &10& \colorbox{gray}{3.701E-3} & 3.752E-3&1.528E-2& \colorbox{red}{2.778E-3} &4.514E-1&& \colorbox{red}{1.156E-3} & \colorbox{gray}{1.615E-3}&6.337E-3&9.203E-1&4.294E-1\\
  && \colorbox{gray}{4.104E-3} & 4.145E-3&1.697E-2& \colorbox{red}{3.235E-3} &5.161E-1&& \colorbox{red}{1.482E-3}& \colorbox{gray}{1.931E-3} &1.076E-1&1.077E+0&4.410E-1\\

  \cline{2-7} \cline{9-13}
  && \colorbox{red}{4.394E-3} & \colorbox{gray}{4.557E-3} &1.360E-2&5.254E-3&5.087E-1&& \colorbox{red}{9.325E-4} & \colorbox{gray}{1.474E-3} &7.110E-3&5.890E-1&4.975E-1\\
  &15& \colorbox{gray}{6.050E-3} & \colorbox{red}{5.863E-3} &1.726E-2&6.005E-3&5.289E-1&& \colorbox{red}{1.517E-3} & \colorbox{gray}{1.881E-3} &3.431E-1&1.133E+0&5.032E-1\\
  && \colorbox{gray}{7.623E-3} & \colorbox{red}{6.929E-3} &2.114E-2&9.409E-3&5.381E-1&& \colorbox{red}{2.427E-3} & \colorbox{gray}{3.159E-3} &1.073E+0&1.249E+0&5.136E-1\\
 \hline
\end{tabular}
}
\end{center}
\end{table*}


\begin{table*}[!htbp]
\caption{Best, Median and Worst HV Values by MOEA/GLU, MOEA/DD,
NSGA-III, MOEA/D and GrEA on DTLZ1, DTLZ2, DTLZ3 and DTLZ4 instances with Different Number of Objectives.
The values in red are the best, and the values in gray are the second best.}
\begin{center}\label{HVonDTLZs}
\resizebox{\textwidth}{!}{ %
\begin{tabular}{|c|c|c|c|c|c|c|c|c|c|c|c|c|}
 \hline
  &m & \centering{\textsf{MOEA/GLU}} & \centering{\textsf{MOEA/DD}}&NSGA-III&MOEA/D&GrEA &&MOEA/GLU&MOEA/DD&NSGA-III&MOEA/D&GrEA\tabularnewline
  \hline

  \multirow{15}{0.2cm}{\begin{sideways}{DTLZ1}\end{sideways}}&& \colorbox{red}{0.973657} & \colorbox{gray}{0.973597} &0.973519&0.973541&0.967404&\multirow{15}{0.2cm}{\begin{sideways}{DTLZ3}\end{sideways}}& \colorbox{red}{0.926717} & \colorbox{gray}{0.926617} &0.926480&0.926598&0.924652 \\
  &3& \colorbox{red}{0.973576} & \colorbox{gray}{0.973510} &0.973217&0.973380&0.964059&& \colorbox{red}{0.926457} & \colorbox{gray}{0.926346}&0.925805&0.925855&0.922650\\
  && \colorbox{red}{0.973279} & \colorbox{gray}{0.973278} &0.971931&0.972484&0.828008&&\colorbox{red}{0.924931} & \colorbox{gray}{0.924901} &0.924234&0.923858&0.621155\\

  \cline{2-7} \cline{9-13}
  && \colorbox{red}{0.998981} & \colorbox{gray}{0.998980} &0.998971& 0.998978 &0.991451&& \colorbox{red}{0.990565} & \colorbox{gray}{0.990558} &0.990453&0.990543&0.963021\\
  &5& \colorbox{red}{0.998976} & \colorbox{gray}{0.998975}&0.998963&0.998969&0.844529&& \colorbox{red}{0.990532} & \colorbox{gray}{0.990515} &0.990344&0.990444&0.808084\\
  && \colorbox{red}{0.998970} & \colorbox{gray}{0.998968} &0.998673&0.998954&0.500179&& \colorbox{red}{0.990451} & \colorbox{gray}{0.990349} &0.989510&0.990258&0.499908\\

  \cline{2-7} \cline{9-13}
  && 0.999948 & \colorbox{gray}{0.999949} & \colorbox{red}{0.999975} &0.999943&0.999144&& \colorbox{red}{0.999345} & \colorbox{gray}{0.999343} &0.999300&0.999328&0.953478\\
  &8& \colorbox{red}{0.999925} & \colorbox{gray}{0.999919} &0.993549&0.999866&0.997992&& \colorbox{red}{0.999322} & \colorbox{gray}{0.999311} &0.924059&0.999303&0.791184\\
  && \colorbox{red}{0.999888} & \colorbox{gray}{0.999887} &0.966432&0.999549&0.902697&& \colorbox{red}{0.999252} & \colorbox{gray}{0.999248} &0.904182&0.508355&0.498580\\

  \cline{2-7} \cline{9-13}
  && \colorbox{gray}{0.999991} & \colorbox{red}{0.999994} & \colorbox{gray}{0.999991} &0.999983&0.999451&& \colorbox{gray}{0.999922}& \colorbox{red}{0.999923} &0.999921&\colorbox{gray}{0.999922}&0.962168\\
  &10& 0.999981 & \colorbox{red}{0.999990} & \colorbox{gray}{0.999985} &0.999979&0.998587&& \colorbox{gray}{0.999921} & \colorbox{red}{0.999922} &0.999918&0.999920&0.735934\\
  && \colorbox{gray}{0.999971} & \colorbox{red}{0.999974} & 0.999969 &0.999956&0.532348&& \colorbox{gray}{0.999919} & \colorbox{red}{0.999921} &0.999910&0.999915&0.499676\\

  \cline{2-7} \cline{9-13}
  && \colorbox{red}{0.999986} & \colorbox{gray}{0.999882} &0.999731&0.999695&0.172492&& \colorbox{red}{0.999996} & \colorbox{gray}{0.999982} &0.999910&0.999918&0.000000\\
  &15& \colorbox{red}{0.999923} & \colorbox{gray}{0.999797} &0.999686&0.999542&0.000000&& \colorbox{red}{0.999994} & \colorbox{gray}{0.999951} &0.999793&0.999792&0.000000\\
  && \colorbox{red}{0.999826}  & \colorbox{gray}{0.999653} &0.999574&0.999333&0.000000&& \colorbox{red}{0.999990} & \colorbox{gray}{0.999915} &0.999780&0.999628&0.000000\\
 \hline
 \hline
   \multirow{15}{0.2cm}{\begin{sideways}{DTLZ2}\end{sideways}}&& \colorbox{red}{0.926698} & \colorbox{gray}{0.926674}&0.926626&0.926666&0.924246&\multirow{15}{0.2cm}{\begin{sideways}{DTLZ4}\end{sideways}}& \colorbox{red}{0.926731} & \colorbox{red}{0.926731} &0.926659& \colorbox{gray}{0.926729} &0.924613 \\
  &3& \colorbox{red}{0.926682} & \colorbox{gray}{0.926653} &0.926536&0.926639&0.923994&& \colorbox{red}{0.926729} & \colorbox{red}{0.926729} &0.926705&\colorbox{gray}{0.926725}&0.924094\\
  && \colorbox{red}{0.926652} & \colorbox{gray}{0.926596} &0.926359&0.926613&0.923675&& \colorbox{red}{0.926725} & \colorbox{red}{0.926725} & \colorbox{gray}{0.799572} &0.500000&0.500000\\

 \cline{2-7} \cline{9-13}
  && \colorbox{red}{0.990545} & \colorbox{gray}{0.990535} &0.990459&0.990529&0.990359&& 0.990570 & \colorbox{gray}{0.990575} &\colorbox{red}{0.991102}&0.990569&0.990514\\
  &5& \colorbox{red}{0.990533} & \colorbox{gray}{0.990527} &0.990400&0.990518&0.990214&& \colorbox{gray}{0.990570} &\colorbox{red}{0.990573} &0.990413&0.990568&0.990409\\
  && \colorbox{red}{0.990513} & \colorbox{gray}{0.990512} &0.990328&0.990511&0.990064&& \colorbox{gray}{0.990569} & \colorbox{red}{0.990570} &0.990156&0.973811&0.990221\\

  \cline{2-7} \cline{9-13}
  && 0.999341 & \colorbox{gray}{0.999346} &0.999320&0.999341& \colorbox{red}{0.999991} && \colorbox{red}{0.999364} & \colorbox{red}{0.999364} & \colorbox{gray}{0.999363} & \colorbox{gray}{0.999363}&0.999102\\
  &8& 0.999326 & \colorbox{gray}{0.999337} &0.978936&0.999329&\colorbox{red}{0.999670} && \colorbox{red}{0.999363} & \colorbox{red}{0.999363} & \colorbox{gray}{0.999361}&0.998497&0.999039\\
  && 0.999296 & \colorbox{red}{0.999329} &0.919680&\colorbox{gray}{0.999307}&0.989264&& \colorbox{red}{0.999362}&0.998360&0.994784&0.995753& \colorbox{gray}{0.998955}\\

  \cline{2-7} \cline{9-13}
  && 0.999921 & \colorbox{red}{0.999952} &0.999918& \colorbox{gray}{0.999922} &0.997636&& \colorbox{gray}{0.999919} & \colorbox{red}{0.999921} &0.999915&0.999918&0.999653\\
  &10& 0.999920 & \colorbox{red}{0.999932} &0.999916& \colorbox{gray}{0.999921} &0.996428&& \colorbox{gray}{0.999914} & \colorbox{red}{0.999920} &0.999910&0.999907&0.999608\\
  && \colorbox{gray}{0.999919} & \colorbox{red}{0.999921} &0.999915& \colorbox{gray}{0.999919}&0.994729&& \colorbox{gray}{0.999910} & \colorbox{red}{0.999917} &0.999827&0.999472&0.999547\\

  \cline{2-7} \cline{9-13}
  && \colorbox{red}{0.999998} & \colorbox{gray}{0.999976} &0.999975&0.999967&0.999524&& \colorbox{red}{0.999990} & \colorbox{gray}{0.999915} &0.999910&0.999813&0.999561\\
  &15& \colorbox{red}{0.999997} & \colorbox{gray}{0.999954} &0.999939&0.999951&0.999496&& \colorbox{red}{0.999979} & \colorbox{gray}{0.999762} &0.999581&0.546405&0.999539\\
  && \colorbox{red}{0.999993} & \colorbox{gray}{0.999915} &0.999887&0.999913&0.998431&& \colorbox{red}{0.999959} & \colorbox{gray}{0.999680} &0.617313&0.502115&0.999521\\
 \hline
\end{tabular}
}
\end{center}
\end{table*}

We calculate the IGD values and HV values of the same solution sets found by MOEA/GLU,
and compare the calculation results with those of MOEA/DD, NSGA-III, MOEA/D and GrEA obtained in\cite{MOEADD}.
\begin{enumerate}
  \item DTLZ1:From the calculation results listed in Table \ref{IGDonDTLZs} and Table \ref{HVonDTLZs},
it can be seen that MOEA/GLU and MOEA/DD perform better than the other three algorithms on all of the IGD values and most of the HV values.
Specifically, MOEA/GLU wins in the best and median IGD values of the 3-, 8-, 10- and 15-objective instances, and
MOEA/DD wins in the worst IGD values of the 3-, 8-, 10- and 15-objective instances.
As for the 5-objective instance, MOEA/GLU wins in all of the IGD values.
When it comes to the HV values, MOEA/GLU performs the best on the 3-, 5- and 15-objective instances,
and MOEA/DD shows the best performance on the 10-objective instance as listed in Table \ref{HVonDTLZs}.
In addition, MOEA/GLU wins in the median and worst HV values of the 8-objective instance, and NSGA-III wins in the best HV value of it.
Although all of the values obtained by MOEA/GLU and MOEA/DD  are close,
MOEA/GLU wins in most of the IGD and HV values. Therefore, MOEA/GLU can be considered as the best optimizer for DTLZ1.
  \item DTLZ2:As it can be seen from Table \ref{IGDonDTLZs}, MOEA/D, MOEA/GLU and MOEA/DD are significantly better than the other two on all of the IGD values of DTLZ2.
  As for the IGD values, MOEA/GLU performs the best on the 3-, 5- and 8-objective instances, and MOEA/D performs the best on the 10-objective instance.
  In addition, MOEA/GLU wins in the best value of the 15-objective instance, and MOEA/DD wins in the median and worst values of it.
  When it comes to the HV values, MOEA/GLU performs the best on the 3-, 5- and 15-objective instances,
  MOEA/DD performs the best on the 10-objective instance and wins in the worst value of the 8-objective instance,
  and GrEA wins in the best and median values of the 8-objective instance.
  On the whole, the differences of MOEA/GLU, MOEA/DD and MOEA/D  are not significant on DTLZ2, but MOEA/GLU wins in more values than both MOEA/D and MOEA/DD.
  Therefore, MOEA/GLU can also be considered as the best optimizer for DTLZ2.
  \item DTLZ3:Again, MOEA/GLU and MOEA/DD are the best two optimizer for DTLZ3, and their performances are also close.
  As for the IGD　values, MOEA/GLU performs the best on the 5- and 8-objective instances,
  MOEA/DD performs the best on the 10-objective instance,
  MOEA/GLU wins in the best and median values of the 3-objective instance and the best value of the 15-objective instance,
  MOEA/DD wins in the median and worst values of the 15-objective instance, and the worst value of the 3-objective.
  As far as the HV values are concerned,
  MOEA/GLU performs the best on the 3-, 5-, 8- and 15-objective instances,
  and MOEA/DD performs the best on the 10-objective instance.
  Since MOEA/GLU wins in more values than the other four algorithms,
  it can be considered as the best optimizer for DTLZ3.
  \item DTLZ4:It is clear that MOEA/GLU performs the best on all of the IGD values of DTLZ4.
  However, it is hard to distinguish the better one from MOEA/GLU and MOEA/DD when it comes to the HV values.
  Interestingly, the performance of MOEA/GLU and MOEA/DD are so close that all of the HV values of the 3-objective instance,
  the best and median HV values of the 8-objective instance obtained by them are equal in terms of 6 significant digits.
  But taking the performance on the IGD values into consideration, MOEA/GLU is  the best optimizer for DTLZ4.
\end{enumerate}

\subsection{Performance Comparisons on WFG1 to WFG9}

\begin{table}[htbp]
    \caption{Best, Median and Worst HV Values by MOEA/GLU, MOEA/DD,
    MOEA/D and GrEA on WFG1 to WFG5 instances with Different Number of Objectives.
    The values in red are the best, and the values in gray are the second best.}
\begin{center}\label{HVonWFGs1To5}
\setlength{\abovecaptionskip}{0pt}
\setlength{\belowcaptionskip}{0pt}
\resizebox{0.49\textwidth}{!}{
\begin{tabular}{|c|c|c|c|c|c|}
\hline
&m&MOEA/GLU&MOEA/DD&MOEA/D&GrEA\\
\hline
  \multirow{12}{0.2cm}{\begin{sideways}{WFG1}\end{sideways}}
  &&\colorbox{gray}{0.937116}&\colorbox{red}{0.937694}&0.932609&0.794748\\
  &3&0.928797&\colorbox{red}{0.933402} &\colorbox{gray}{0.929839}&0.692567\\
  &&\colorbox{red}{0.915136}&\colorbox{gray}{0.899253} &0.815356&0.627963\\

  \cline{2-6}
  &&0.906874&\colorbox{red}{0.963464} &\colorbox{gray}{0.918652}&0.876644\\
  &5&0.899351&\colorbox{red}{0.960897} &\colorbox{gray}{0.915737}&0.831814\\
  &&0.862874 &\colorbox{red}{0.959840} &\colorbox{gray}{0.912213}&0.790367\\

  \cline{2-6}
  &&0.839662 &\colorbox{red}{0.922284}&\colorbox{gray}{0.918252}&0.811760\\
  &8&0.831208  &\colorbox{red}{0.913024} &\colorbox{gray}{0.911586}&0.681959\\
  &&0.781919  &\colorbox{red}{0.877784} &\colorbox{gray}{0.808931}&0.616006\\

  \cline{2-6}
  && 0.887565 &\colorbox{red}{0.926815}  &\colorbox{gray}{0.922484}&0.866298\\
  &10& 0.843225 &\colorbox{red}{0.919789} &\colorbox{gray}{0.915715}&0.832016\\
  &&0.794202  &\colorbox{red}{0.864689} &\colorbox{gray}{0.813928}&0.757841\\

 \hline
   \hline

  \multirow{12}{0.2cm}{\begin{sideways}{WFG2}\end{sideways}}
  &&\colorbox{red}{0.959834}&\colorbox{gray}{0.958287}&0.951685&0.950084 \\
  &3&\colorbox{red}{0.958155}&\colorbox{gray}{0.952467}&0.803246&0.942908\\
  &&\colorbox{red}{0.808454}&\colorbox{gray}{0.803397}&0.796567&0.800186\\

  \cline{2-6}
  &&\colorbox{red}{0.995169}&\colorbox{gray}{0.986572}&0.982796&0.980806\\
  &5&\colorbox{red}{0.993049}&\colorbox{gray}{0.985129}&0.978832&0.976837\\
  &&\colorbox{gray}{0.813859}&\colorbox{red}{0.980035}&0.807951&0.808125\\

  \cline{2-6}
  &&0.978775&\colorbox{red}{0.981673}&0.963691&\colorbox{gray}{0.980012}\\
  &8&0.795215&\colorbox{red}{0.967265}&0.800333&\colorbox{gray}{0.840293}\\
  &&\colorbox{gray}{0.778920}&\colorbox{red}{0.789739}&0.787271&0.778291\\

  \cline{2-6}
  &&\colorbox{red}{0.981398}&\colorbox{gray}{0.968201}&0.962841&0.964235\\
  &10&\colorbox{red}{0.978021}&\colorbox{gray}{0.965345}&0.957434&0.959740\\
  &&0.779176&\colorbox{red}{0.961400}&0.773474&\colorbox{gray}{0.956533}\\

 \hline
 \hline

 \multirow{12}{0.2cm}{\begin{sideways}{WFG3}\end{sideways}}&&\colorbox{gray}{0.700589}
  &\colorbox{red}{0.703664}&0.697968&0.699502 \\
  &3&\colorbox{gray}{0.695748}&\colorbox{red}{0.702964}&0.692355&0.672221\\
  &&\colorbox{gray}{0.689587}&\colorbox{red}{0.701624}&0.679281&0.662046\\

  \cline{2-6}
  &&\colorbox{gray}{0.679497}&0.673031&0.669009&\colorbox{red}{0.695221}\\
  &5&\colorbox{gray}{0.675726}&0.668938&0.662925&\colorbox{red}{0.684583}\\
  &&0.662165&\colorbox{gray}{0.662951}&0.654729&\colorbox{red}{0.671553}\\

  \cline{2-6}
  &&0.572932&\colorbox{gray}{0.598892}&0.529698&\colorbox{red}{0.657744}\\
  &8&0.554256&\colorbox{gray}{0.565609}&0.457703&\colorbox{red}{0.649020}\\
  &&0.526689&\colorbox{gray}{0.556725}&0.439274&\colorbox{red}{0.638147}\\

  \cline{2-6}
  &&\colorbox{red}{0.572593}&\colorbox{gray}{0.552713}&0.382068&0.543352\\
  &10&\colorbox{red}{0.554042}&\colorbox{gray}{0.532897}&0.337978&0.513261\\
  &&\colorbox{red}{0.531208}&\colorbox{gray}{0.504943}&0.262496&0.501210\\

 \hline
 \hline

  \multirow{12}{0.2cm}{\begin{sideways}{WFG4}\end{sideways}}
  &&\colorbox{red}{0.731535}&\colorbox{gray}{0.727060}&0.724682&0.723403\\
  &3&\colorbox{red}{0.731180}  &\colorbox{gray}{0.726927} &0.723945&0.722997\\
  &&\colorbox{red}{0.730558}  &\colorbox{gray}{0.726700} &0.723219&0.722629\\

   \cline{2-6}
  &&\colorbox{red}{0.883419}  &0.876181 &0.870868&\colorbox{gray}{0.881161}\\
  &5&\colorbox{red}{0.881701}  &0.875836 &0.862132&\colorbox{gray}{0.879484}\\
  &&\colorbox{red}{0.880210}  &0.875517 &0.844219&\colorbox{gray}{0.877642}\\

  \cline{2-6}
  &&\colorbox{red}{0.939271}  &\colorbox{gray}{0.920869}  &0.784340&0.787287\\
  &8&\colorbox{red}{0.933853}  &\colorbox{gray}{0.910146} &0.737386&0.784141\\
  &&\colorbox{red}{0.926261}  &\colorbox{gray}{0.902710} &0.718648&0.679178\\

  \cline{2-6}
  &&\colorbox{red}{0.967623}  &\colorbox{gray}{0.913018}  &0.747485&0.896261\\
   &10&\colorbox{red}{0.963674}  &\colorbox{gray}{0.907040} &0.712680&0.843257\\
  &&\colorbox{red}{0.951068}  &\colorbox{gray}{0.888885} &0.649713&0.840257\\

 \hline
 \hline
  \multirow{12}{0.2cm}{\begin{sideways}{WFG5}\end{sideways}}
  &&\colorbox{red}{0.698469}&\colorbox{gray}{0.693665}&0.693135&0.689784\\
  &3&\colorbox{gray}{0.692607}&\colorbox{red}{0.693544}&0.687378&0.689177\\
  &&0.685518&\colorbox{red}{0.691173}&0.681305&\colorbox{gray}{0.688885}\\

   \cline{2-6}
  &&\colorbox{red}{0.844325}&0.833159&0.829696&\colorbox{gray}{0.836232}\\
  &5&\colorbox{red}{0.841781}&0.832710&0.826739&\colorbox{gray}{0.834726}\\
  &&\colorbox{red}{0.838402}&0.830367&0.812225&\colorbox{gray}{0.832212}\\

  \cline{2-6}
  &&\colorbox{red}{0.892830}&\colorbox{gray}{0.852838}&0.779091&0.838183\\
  &8&\colorbox{red}{0.889458}&\colorbox{gray}{0.846736}&0.753486&0.641973\\
  &&\colorbox{red}{0.884971}&\colorbox{gray}{0.830338}&0.705938&0.571933\\

  \cline{2-6}
  &&\colorbox{red}{0.919163}&\colorbox{gray}{0.848321}&0.730990&0.791725\\
  &10&\colorbox{red}{0.916148}&\colorbox{gray}{0.841118}&0.715161&0.725198\\
  &&\colorbox{red}{0.911875}&\colorbox{gray}{0.829547}&0.673789&0.685882\\
 \hline

\end{tabular}
}
\end{center}
\end{table}

\begin{table}[htbp]
    \caption{Best, Median and Worst HV Values by MOEA/GLU, MOEA/DD,
    MOEA/D and GrEA on WFG6 to WFG9 instances with Different Number of Objectives.
    The values in red are the best, and the values in gray are the second best.}
\begin{center}\label{HVonWFG6toWFG9}
\setlength{\abovecaptionskip}{0pt}
\setlength{\belowcaptionskip}{0pt}
\resizebox{0.49\textwidth}{!}{
\begin{tabular}{|c|c|c|c|c|c|}
    \hline
    &m&MOEA/GLU&MOEA/DD&MOEA/D&GrEA\\
    \hline
  \multirow{12}{0.2cm}{\begin{sideways}{WFG6}\end{sideways}}
  &&\colorbox{red}{0.710228}&\colorbox{gray}{0.708910}&0.702840&0.699876\\
&3&\colorbox{red}{0.701988}&\colorbox{gray}{0.699663}&0.695081&0.693984\\
  &&\colorbox{red}{0.698358}&\colorbox{gray}{0.689125}&0.684334&0.685599\\

   \cline{2-6}
  &&\colorbox{red}{0.858096}&0.850531&0.846015&\colorbox{gray}{0.855839}\\
  &5&\colorbox{gray}{0.846655}&0.838329&0.813844&\colorbox{red}{0.847137}\\
  &&\colorbox{gray}{0.840335}&0.828315&0.754054&\colorbox{red}{0.840637}\\

  \cline{2-6}
  &&\colorbox{red}{0.912150}&0.876310&0.692409&\colorbox{gray}{0.912095}\\
  &8&\colorbox{gray}{0.901300}&0.863087&0.661156&\colorbox{red}{0.902638}\\
  &&\colorbox{gray}{0.880581}&0.844535&0.567108&\colorbox{red}{0.885712}\\

  \cline{2-6}
  &&\colorbox{gray}{0.938343}&0.884394&0.643198&\colorbox{red}{0.943454}\\
  &10&\colorbox{red}{0.927854}&0.859986&0.582342&\colorbox{gray}{0.927443}\\
  &&\colorbox{red}{0.914464}&0.832299&0.409210&\colorbox{gray}{0.884145}\\

\hline
\hline

\multirow{12}{0.2cm}{\begin{sideways}{WFG7}\end{sideways}}
  &&\colorbox{red}{0.731908}&\colorbox{gray}{0.727069}&0.725252&0.723229\\
  &3&\colorbox{red}{0.731809}  &\colorbox{gray}{0.727012} &0.724517&0.722843\\
  &&\colorbox{red}{0.731691}  &\colorbox{gray}{0.726907} &0.723449&0.722524\\

   \cline{2-6}
  &&\colorbox{red}{0.888158}  &0.876409 &0.859727&\colorbox{gray}{0.884174}\\
  &5&\colorbox{red}{0.887856}  &0.876297 &0.843424&\colorbox{gray}{0.883079}\\
  &&\colorbox{red}{0.887592}  &0.874909 &0.811292&\colorbox{gray}{0.881305}\\

  \cline{2-6}
  &&\colorbox{red}{0.948854}  &\colorbox{gray}{0.920763}  &0.729953&0.918742\\
  &8&\colorbox{red}{0.947862}  &\colorbox{gray}{0.917584} &0.708701&0.910023\\
  &&\colorbox{red}{0.946082}  &\colorbox{gray}{0.906219} &0.605900&0.901292\\

  \cline{2-6}
  &&\colorbox{red}{0.976171}  &0.927666  &0.706473&\colorbox{gray}{0.937582}\\
  &10&\colorbox{red}{0.975644}  &\colorbox{gray}{0.923441} &0.625828&0.902343\\
  &&\colorbox{red}{0.974641}  &\colorbox{gray}{0.917141} &0.596189&0.901477\\

  \hline
  \hline

  \multirow{12}{0.2cm}{\begin{sideways}{WFG8}\end{sideways}}
  &&\colorbox{red}{0.678825}&\colorbox{gray}{0.672022}&0.671355&0.671845\\
  &3&\colorbox{red}{0.677146}&\colorbox{gray}{0.670558}&0.669927&0.669762\\
  &&\colorbox{red}{0.674987}&\colorbox{gray}{0.668593}&0.664120&0.667948\\

  \cline{2-6}
  &&0.806626&\colorbox{red}{0.818663}&\colorbox{gray}{0.808204}&0.797496\\
  &5&\colorbox{red}{0.805050}&\colorbox{gray}{0.795215}&0.793773&0.792692\\
  &&\colorbox{red}{0.803366}&\colorbox{gray}{0.792900}&0.771763&0.790693\\

  \cline{2-6}
  &&\colorbox{red}{0.895652}&\colorbox{gray}{0.876929}&0.537772&0.803050\\
  &8&\colorbox{gray}{0.845761}&\colorbox{red}{0.845975}&0.446544&0.799986\\
  &&\colorbox{red}{0.823666}&0.730348&0.347990&\colorbox{gray}{0.775434}\\

  \cline{2-6}
  &&\colorbox{red}{0.961919}&\colorbox{gray}{0.896317}&0.508652&0.841704\\
  &10&\colorbox{red}{0.923244}&\colorbox{gray}{0.844036}&0.350409&0.838256\\
  &&\colorbox{red}{0.881384}&0.715250&0.270931&\colorbox{gray}{0.830394}\\

  \hline
  \multirow{12}{0.2cm}{\begin{sideways}{WFG9}\end{sideways}}
  &&0.695369&\colorbox{red}{0.707269}&0.688940&\colorbox{gray}{0.702489}\\
  &3&0.642755&\colorbox{red}{0.687401}&\colorbox{gray}{0.681725}&0.638103\\
  &&\colorbox{red}{0.642240}&\colorbox{gray}{0.638194}&0.636355&0.636575\\

  \cline{2-6}
  &&0.809717&\colorbox{red}{0.834616}&0.798069&\colorbox{gray}{0.823916}\\
  &5&0.751592&\colorbox{red}{0.797185}&\colorbox{gray}{0.789998}&0.753683\\
  &&\colorbox{gray}{0.749481}&\colorbox{red}{0.764723}&0.727728&0.747315\\

  \cline{2-6}
  &&\colorbox{gray}{0.828505}&0.772671&0.633476&\colorbox{red}{0.842953}\\
  &8&\colorbox{gray}{0.809564}&0.759369&0.604016&\colorbox{red}{0.831775}\\
  &&\colorbox{gray}{0.746497}&0.689923&0.548119&\colorbox{red}{0.765730}\\

  \cline{2-6}
  &&\colorbox{gray}{0.843321}&0.717168&0.572925&\colorbox{red}{0.860676}\\
  &10&\colorbox{red}{0.830062}&\colorbox{gray}{0.717081}&0.546451&0.706632\\
  &&\colorbox{red}{0.803744}&\colorbox{gray}{0.696061}&0.516309&0.686917\\
\hline

\end{tabular}
}
\end{center}
\end{table}

The HV values of MOEA/GLU, MOEA/DD, MOEA/D and GrEA on WFG1 to WFG5 are listed in Table \ref{HVonWFGs1To5}, and the HV values on WFG6 to WFG9 are listed in Table \ref{HVonWFG6toWFG9}. The comparison results can be concluded as follows.
\begin{enumerate}
  \item WFG1:MOEA/DD wins in all the values of WFG1 except the worst value of the 3-objective instance, and hence be regarded as the  best optimizer for WFG1.
  \item WFG2:MOEA/GLU shows the best performance on the 3-objective instance, while MOEA/DD performs the best on the 8-objective instance.
  In addition, MOEA/GLU wins in the best and median values of 5- and 10-objective instances, while MOEA/DD wins in the worst values  of them.
  Obviously, MOEA/GLU and MOEA/DD are the best two optimizer for WFG2, but it is hard to tell which one is better, since the differences between them are not significant.
  \item WFG3:MOEA/GLU performs the best on the 10-objective instance, MOEA/DD shows the best performance on the 3-objective instance, and GrEA wins in
  the 5- and 8-objective instances. The values obtained by the three algorithms are very close. They all have their own advantages.
  \item WFG4:MOEA/GLU shows the best in all values of WFG4, and is considered as the winner.
  \item WFG5:Like in WFG4, MOEA/GLU is the winner of WFG5,
  since it wins in all values except the median and worst values of the 3-objective instance.
  \item WFG6:MOEA/GLU and GrEA are the best two optimizer of WFG.
  The values obtained by them are not significant with ups and downs on both sides.
  Specifically, MOEA/GLU wins in  the 3-objective instance, the best values of the 5- and 8-objective
  instances, the median and worst values of the 10-objective instance. GrEA wins in all the other values.
  \item WFG7:MOEA/GLU wins in all the values of WFG7, and is considered as the best optimizer.
  \item WFG8:MOEA/GLU wins in most of the values of WFG8 except the best value of the 5-objective instance
  and the median value of the 8-objective instance. Therefore, it can also be regarded as the best optimizer for WFG8.
  \item WFG9:The situation of WFG9 is a little bit complicated, but it is clear that MOEA/GLU, MOEA/DD and GrEA are all better than MOEA/D.
  To be specific, GrEA wins in the 8-objective instance, and it might be said that
  MOEA/DD performs the best on the 3- and 5-objective instance  although the worst value of it on the 3-objective instance is slightly  worse than that of MOEA/GLU.
  In addition, the median and worst values of MOEA/GLU  on the 10-objective instance are far better than those of other algorithms, while the best value is sightly worse than that of GrEA.
\end{enumerate}

On the whole, MOEA/GLU shows a very competitive performance on the WFG test suite, especially  WFG4, WFG5, WFG7 and WFG8, of which MOEA/GLU wins almost all the HV values.



\subsection{Performance Comparisons of Algorithms with Different Criteria for Comparison}

\begin{table*}[htbp]
\caption{Best, Median and Worst HV Values by MOEA/GLU with Three Different Criteria:
PBI, H1 and H2 on  instances of WFG1 to WFG9 with 3,5,8 and 10 Objectives.
The values in red font are the best.}
\begin{center}
\setlength{\abovecaptionskip}{0pt}
\setlength{\belowcaptionskip}{0pt}
\resizebox{\textwidth}{!}{ %
\begin{tabular}{|c|c|c|c|c|c|c|c|c|c|c|c|c|c}
 \hline
 &m & PBI & H1&H2&&PBI&H1&H2&&PBI&H1&H2\tabularnewline
  \hline
  \multirow{12}{0.2cm}{\begin{sideways}{WFG1}\end{sideways}}&&0.932478&0.937488&\emph{\textcolor[rgb]{1,0,0}{0.944690}}&\multirow{12}{0.2cm}{\begin{sideways}{WFG2}
  \end{sideways}}&0.957614&\emph{\textcolor[rgb]{1,0,0}{0.958855}}&0.958309&\multirow{12}{0.2cm}{\begin{sideways}{WFG3}\end{sideways}}&0.690390&\emph{\textcolor[rgb]{1,0,0}{0.700632}}&0.699646 \\
  &3&0.919064&0.924179&\emph{\textcolor[rgb]{1,0,0}{0.939554}}&&0.955452&0.884332&\emph{\textcolor[rgb]{1,0,0}{0.955588}}&&0.685735&\emph{\textcolor[rgb]{1,0,0}{0.696145}}&0.692045\\
  & &0.907255&0.903350&\emph{\textcolor[rgb]{1,0,0}{0.922585}}&&0.810162&0.810262&\emph{\textcolor[rgb]{1,0,0}{0.811369}}&&0.675661&0.684853&\emph{\textcolor[rgb]{1,0,0}{0.685778}}\\

   \cline{2-5} \cline{7-9}\cline{11-13}
  & &0.909324&0.908070&\emph{\textcolor[rgb]{1,0,0}{0.931487}}&&0.994983&0.994851&\emph{\textcolor[rgb]{1,0,0}{0.997409}}&&0.679563&0.681048&\emph{\textcolor[rgb]{1,0,0}{0.682644}}\\
  &5&0.897364 &0.898754 &\emph{\textcolor[rgb]{1,0,0}{0.922682}}&&0.993170&0.993575&\emph{\textcolor[rgb]{1,0,0}{0.996668}}&&0.673978&0.676910&\emph{\textcolor[rgb]{1,0,0}{0.676994}}\\
  & &0.860576 &0.873925  &\emph{\textcolor[rgb]{1,0,0}{0.895859}}&&0.814648&0.815756&\emph{\textcolor[rgb]{1,0,0}{0.816768}}&&\emph{\textcolor[rgb]{1,0,0}{0.669298}}&0.666466&0.666298\\

   \cline{2-5} \cline{7-9}\cline{11-13}
  & &0.843381&0.844850&\emph{\textcolor[rgb]{1,0,0}{0.918863}}&&0.978952&0.976819&\emph{\textcolor[rgb]{1,0,0}{0.992986}}&&0.586041&0.578472&\emph{\textcolor[rgb]{1,0,0}{0.605893}}\\
  &8&0.831511&0.787144&\emph{\textcolor[rgb]{1,0,0}{0.867385}}&&0.961105&0.970539&\emph{\textcolor[rgb]{1,0,0}{0.983890}}&&0.550607&0.552667&\emph{\textcolor[rgb]{1,0,0}{0.592805}}\\
  & &0.743512&0.706517&\emph{\textcolor[rgb]{1,0,0}{0.851531}}&&0.776725&0.773388&\emph{\textcolor[rgb]{1,0,0}{0.800754}}&&0.532548&0.540043&\emph{\textcolor[rgb]{1,0,0}{0.581368}}\\

   \cline{2-5} \cline{7-9}\cline{11-13}
  & &0.849143&0.876852&\emph{\textcolor[rgb]{1,0,0}{0.918984}}&&0.982992&0.980088&\emph{\textcolor[rgb]{1,0,0}{0.995465}}&&\emph{\textcolor[rgb]{1,0,0}{0.574988}}&0.565746&0.574100\\
  &10&0.844050&0.846479&\emph{\textcolor[rgb]{1,0,0}{0.877223}}&&0.978054&0.788371&\emph{\textcolor[rgb]{1,0,0}{0.993171}}&&0.549540&0.548960&\emph{\textcolor[rgb]{1,0,0}{0.555640}}\\
  & &0.785389 &0.746061  &\emph{\textcolor[rgb]{1,0,0}{0.869973}}&&0.781965&0.770895&\emph{\textcolor[rgb]{1,0,0}{0.801428}}&&0.529822&0.508638&\emph{\textcolor[rgb]{1,0,0}{0.534696}}\\

 \hline
 \hline

 \multirow{12}{0.2cm}{\begin{sideways}{WFG4}\end{sideways}}&&0.727187&\emph{\textcolor[rgb]{1,0,0}{0.731777}}&0.731411&\multirow{12}{0.2cm}{\begin{sideways}{WFG5}
  \end{sideways}}&0.697911&\emph{\textcolor[rgb]{1,0,0}{0.698434}}&0.698400&\multirow{12}{0.2cm}{\begin{sideways}{WFG6}\end{sideways}}&0.708928&\emph{\textcolor[rgb]{1,0,0}{0.713367}}&0.713228 \\
  &3&0.726180&\emph{\textcolor[rgb]{1,0,0}{0.731550}}&0.730869&&\emph{\textcolor[rgb]{1,0,0}{0.695548}}&0.695167&0.693015&&0.699913&\emph{\textcolor[rgb]{1,0,0}{0.703288}}&0.702914\\
  & &0.724841&\emph{\textcolor[rgb]{1,0,0}{0.731255}}&0.730343&&0.690822&0.685562&\emph{\textcolor[rgb]{1,0,0}{0.692078}}&&0.692277&\emph{\textcolor[rgb]{1,0,0}{0.699125}}&0.694282\\

   \cline{2-5} \cline{7-9}\cline{11-13}
  & &0.881501&\emph{\textcolor[rgb]{1,0,0}{0.885457}}&0.883788&&0.843774&\emph{\textcolor[rgb]{1,0,0}{0.845047}}&0.844654&&\emph{\textcolor[rgb]{1,0,0}{0.859456}}&0.856809&0.858598\\
  &5&0.879884&\emph{\textcolor[rgb]{1,0,0}{0.884489}}&0.881128&&0.840175&\emph{\textcolor[rgb]{1,0,0}{0.844616}}&0.842033&&0.847747&\emph{\textcolor[rgb]{1,0,0}{0.849112}}&0.848262\\
  & &0.876368&\emph{\textcolor[rgb]{1,0,0}{0.882801}}&0.878642&&0.833143&\emph{\textcolor[rgb]{1,0,0}{0.843364}}&0.838127&&0.832130&\emph{\textcolor[rgb]{1,0,0}{0.840610}}&0.838766\\

   \cline{2-5} \cline{7-9}\cline{11-13}
  & &0.934740&\emph{\textcolor[rgb]{1,0,0}{0.940762}}&0.938517&&0.891617&\emph{\textcolor[rgb]{1,0,0}{0.895639}}&0.892399&&0.916292&0.913827&\emph{\textcolor[rgb]{1,0,0}{0.916850}}\\
  &8&0.930307&\emph{\textcolor[rgb]{1,0,0}{0.937541}}&0.928525&&0.888071&\emph{\textcolor[rgb]{1,0,0}{0.893686}}&0.888761&&0.900157&\emph{\textcolor[rgb]{1,0,0}{0.902256}}&0.897580\\
  & &0.921402&\emph{\textcolor[rgb]{1,0,0}{0.930123}}&0.920050&&0.882510&\emph{\textcolor[rgb]{1,0,0}{0.887629}}&0.882556&&0.883341&\emph{\textcolor[rgb]{1,0,0}{0.893311}}&0.884989\\

   \cline{2-5} \cline{7-9}\cline{11-13}
  & &0.967500&\emph{\textcolor[rgb]{1,0,0}{0.971182}}&0.963828&&0.917412&\emph{\textcolor[rgb]{1,0,0}{0.920249}}&0.917481&&0.937329&\emph{\textcolor[rgb]{1,0,0}{0.941969}}&0.938397\\
  &10&0.961761&\emph{\textcolor[rgb]{1,0,0}{0.968851}}&0.959950&&0.916424&\emph{\textcolor[rgb]{1,0,0}{0.918867}}&0.915163&&0.923721&\emph{\textcolor[rgb]{1,0,0}{0.928654}}&0.926283\\
  & &0.951755&\emph{\textcolor[rgb]{1,0,0}{0.964127}}&0.953305&&0.909849&\emph{\textcolor[rgb]{1,0,0}{0.916739}}&0.911581&&0.911045&0.915935&\emph{\textcolor[rgb]{1,0,0}{0.916056}}\\

 \hline
\hline

 \multirow{12}{0.2cm}{\begin{sideways}{WFG7}\end{sideways}}&&0.729737&\emph{\textcolor[rgb]{1,0,0}{0.731874}}&0.731854&\multirow{12}{0.2cm}{\begin{sideways}{WFG8}
  \end{sideways}}&0.672998&\emph{\textcolor[rgb]{1,0,0}{0.678718}}&0.677100&\multirow{12}{0.2cm}{\begin{sideways}{WFG9}
  \end{sideways}}&0.689751&\emph{\textcolor[rgb]{1,0,0}{0.697593}}&0.693049 \\
  &3&0.728050&\emph{\textcolor[rgb]{1,0,0}{0.731773}}&0.731742&&0.671628&\emph{\textcolor[rgb]{1,0,0}{0.676451}}&0.675703&&0.641612&\emph{\textcolor[rgb]{1,0,0}{0.642693}}&0.642552\\
  & &0.727450&0.731506&\emph{\textcolor[rgb]{1,0,0}{0.731573}}&&0.666242&\emph{\textcolor[rgb]{1,0,0}{0.674706}}&0.671803&&0.638641&\emph{\textcolor[rgb]{1,0,0}{0.642119}}&0.641827\\

   \cline{2-5} \cline{7-9}\cline{11-13}
  & &0.887632&\emph{\textcolor[rgb]{1,0,0}{0.888112}}&0.887789&&0.805790&0.808181&\emph{\textcolor[rgb]{1,0,0}{0.821287}}&&\emph{\textcolor[rgb]{1,0,0}{0.806133}}&0.799793&0.790106\\
  &5&0.887157&\emph{\textcolor[rgb]{1,0,0}{0.888013}}&0.887662&&0.804308&\emph{\textcolor[rgb]{1,0,0}{0.805537}}&0.804396&&0.750934&\emph{\textcolor[rgb]{1,0,0}{0.753767}}&0.750096\\
  & &0.886747&\emph{\textcolor[rgb]{1,0,0}{0.887752}}&0.887201&&0.803095&\emph{\textcolor[rgb]{1,0,0}{0.804091}}&0.801026&&0.748504&\emph{\textcolor[rgb]{1,0,0}{0.749479}}&0.746307\\

   \cline{2-5} \cline{7-9}\cline{11-13}
  & &0.946957&\emph{\textcolor[rgb]{1,0,0}{0.948562}}&0.947672&&0.877491&0.889190&\emph{\textcolor[rgb]{1,0,0}{0.902643}}&&0.816273&\emph{\textcolor[rgb]{1,0,0}{0.824253}}&0.823161\\
  &8&0.945634&\emph{\textcolor[rgb]{1,0,0}{0.947616}}&0.946120&&0.844134&\emph{\textcolor[rgb]{1,0,0}{0.844953}}&0.842148&&0.803022&\emph{\textcolor[rgb]{1,0,0}{0.803785}}&0.797496\\
  & &0.944550&\emph{\textcolor[rgb]{1,0,0}{0.945989}}&0.944323&&0.811508&\emph{\textcolor[rgb]{1,0,0}{0.8248850}}&0.819766&&\emph{\textcolor[rgb]{1,0,0}{0.761689}}&0.740563&0.741695\\

   \cline{2-5} \cline{7-9}\cline{11-13}
  & &0.976190&\emph{\textcolor[rgb]{1,0,0}{0.976717}}&0.976109&&0.942152&0.934571&\emph{\textcolor[rgb]{1,0,0}{0.948360}}&&0.837137&\emph{\textcolor[rgb]{1,0,0}{0.839769}}&0.835042\\
  &10&0.975073&\emph{\textcolor[rgb]{1,0,0}{0.975950}}&0.975063&&0.905199&0.908858&\emph{\textcolor[rgb]{1,0,0}{0.924340}}&&\emph{\textcolor[rgb]{1,0,0}{0.825754}}&0.825240&0.823727\\
  & &0.974186&\emph{\textcolor[rgb]{1,0,0}{0.975258}}&0.974119&&0.877100&0.885137&\emph{\textcolor[rgb]{1,0,0}{0.886750}}&&0.762914&0.757464&\emph{\textcolor[rgb]{1,0,0}{0.764740}}\\

 \hline

\end{tabular}
}
\end{center}\label{HVsofDiffCriteria}
\end{table*}

In this subsection, we compare the HV values of  MOEA/GLU with different criteria for comparing solutions on WFG1 to WFG9 with different objectives.
The HV values are listed in Table \ref{HVsofDiffCriteria}.
The comparison results can be concluded as follows.
For the sake of convenience, we denote MOEA/GLU with the PBI, H1, and H2 criteria as MOEA/GLU-PBI, MOEA/GLU-H1, and MOEA/GLU-H2, respectively.
\begin{itemize}
  \item WFG1:MOEA/GLU-H2 is the best optimizer since it performs the best on all instances of WFG1.
  \item WFG2:MOEA/GLU-H2 wins in all values of WFG2 except the best value of the 3-objective instance.
  Therefore, it is considered to be the best optimizer for WFG2.
  \item WFG3:The situation is a little bit complicated for WFG3. Specifically, MOEA/GLU-PBI wins on the worst value of the 5-objective  instance
  and the best value of the 8-objective instance. And MOEA/GLU-H1 wins in the best and median values of the 3-objective instance,
  while MOEA/GLU-H2 is the best on other values. Therefore, MOEA/GLU-H2 can be considered the best optimizer for WFG3.
  \item WFG4: MOEA/GLU-H1 is the best optimizer that wins in all the values of WFG4.
  \item WFG5: MOEA/GLU-PBI has the best median value for the 3-objective instance of WFG5,
  and MOEA/GLU-H2 wins in the worst value of the 3-objective
  instance, while MOEA/GLU-H1 performs the best on all the other values. Therefore, MOEA/GLU-H1 is considered the best optimizer for WFG5.
  \item WFG6: MOEA/GLU-PBI has the best median value for the 5-objective instance of WFG6,
  and MOEA/GLU-H2 wins in the best value of the 8-objective instance  and the worst value of the 10-objective instance,
  while MOEA/GLU-H1 performs the best on all other values. Therefore, MOEA/GLU-H1 is considered  the best optimizer for WFG6.
  \item WFG7: Since MOEA/GLU-H1 wins in all the values of WFG7 except the worst value of its 3-objective instance,
  it is considered the best optimizer.
  \item WFG8: MOEA/GLU-PBI performs the worst on all the values of WFG8.
  MOEA/GLU-H1 wins in its 3-objective instance, while MOEA/GLU-H2 wins in the 10-objective instance.
  As for 5- and 8-objective  instance,  MOEA/GLU-H1 wins in the median and worst value, and MOEA/GLU-H2 wins in the best value.
  It is clear that MOEA/GLU-PBI is the worst optimizer for WFG8.
  However, as for MOEA/GLU-H1 and MOEA/GLU-H2, it is still hard to say which one of the two is better for WFG8.
  \item WFG9: MOEA/GLU-PBI wins in the best value of the 5-objective instance, the worst value of the 8-objective instance ,
  and the median value of 10-objective instance of WFG9.
  MOEA/GLU-H2 only wins in the worst value of the 10-objective instance, while MOEA/GLU-H1 wins in all the other values of WFG9.
  Therefore, MOEA/GLU-H1 can be considered the best optimizer for WFG9.
\end{itemize}

On the whole, MOEA/GLU-H1 is the best optimizer for WFG4 to WFG7, and MOEA/GLU-H2 is the best for WFG1 to WFG3, and WFG9.
As for WFG8, Both MOEA/GLU-H1 and MOEA/GLU-H2 are better than MOEA/GLU-PBI, but it is hard to say which one of the two is better.
These indicate that the running results of MOEA/GLU  are affected by the criterion for comparing solutions it adopts.

\section{Conclusion}
In this paper, we propose a MOEA with the so-called GLU strategy, i.e., MOEA/GLU.
The main ideas of MOEA/GLU can be concluded as follows.
Firstly, MOEA/GLU employs a set of weight vectors to decompose a given MOP into a set of subproblems
and optimizes them simultaneously, which is similar to other decomposition-based MOEAs.
Secondly, each individual is attached to a weight vector and a weight vector owns only
one individual in MOEA/GLU, which is the same as that in MOEA/D, but different from that in MOEA/DD.
Thirdly, MOEA/GLU adopts a global update strategy, i.e. the GLU strategy.
Our experiments indicate that the GLU strategy can overcome the disadvantages of MOEAs with local update strategies discussed in section II, although it makes the time complexity of the algorithm higher than that of MOEA/D.
These three main ideas make MOEA/GLU a different algorithm from other MOEAs, such as MOEA/D, MOEA/DD, and NSGA-III, etc.
Additionally, the GLU strategy is simpler than the update strategies of MOEA/DD and NSGA-III.
And the time complexity of MOEA/GLU is the same as that of MOEA/DD, but worse than that of MOEA/D.

Our algorithm is compared to several other MOEAs, i.e., MOEA/D, MOEA/DD, NSGA-III, GrEA on 3, 5, 8, 10, 15-objective instances of DTLZ1 to DTLZ4, and
3, 5, 8, 10-objective instances of WFG1 to WFG9.
The experimental results show that our algorithm wins in most of the instances.
In addition, we suggest two hybrid criteria for comparing solutions, and compare them with the PBI criterion.
The empirical results show that the two hybrid criteria is very competitive in  3, 5, 8, 10-objective instances of WFG1 to WFG9.

Our future work can be carried out in the following three aspects.
Firstly, it is interesting to study the performances of MOEA/GLU on other MOPs,
such as the ZDT test problems, the CEC2009 test problems,
combinatorial optimization problems appeared in\cite{Zitzler1999Multiobjective,Ishibuchi2010Many},
and especially some real-world problems with a large number of objectives.
Secondly, it might be valuable to apply the two hybrid criteria for comparing solutions to other MOEAs.
Thirdly, improve MOEA/GLU to overcome its shortcomings.
As we can see, the algorithm  contains at least two shortcomings.
One is that all of its experimental results on WFG1 are worse than those of MOEA/DD except for the best HV value of the 3-objective instance.
The other is that its time complexity is worse than that of MOEA/D.
Further research is necessary to be carried out to try to overcome these two shortcomings.



\section*{Acknowledgment}

The authors would like to thank Qingfu Zhang and Ke Li for
their generously giving the java codes of MOEA/D and MOEA/DD.

\ifCLASSOPTIONcaptionsoff
  \newpage
\fi



\bibliographystyle{IEEEtran}
\bibliography{moea}
%



%

\begin{IEEEbiography}[{\includegraphics[width=1in,height=1.25in,clip]{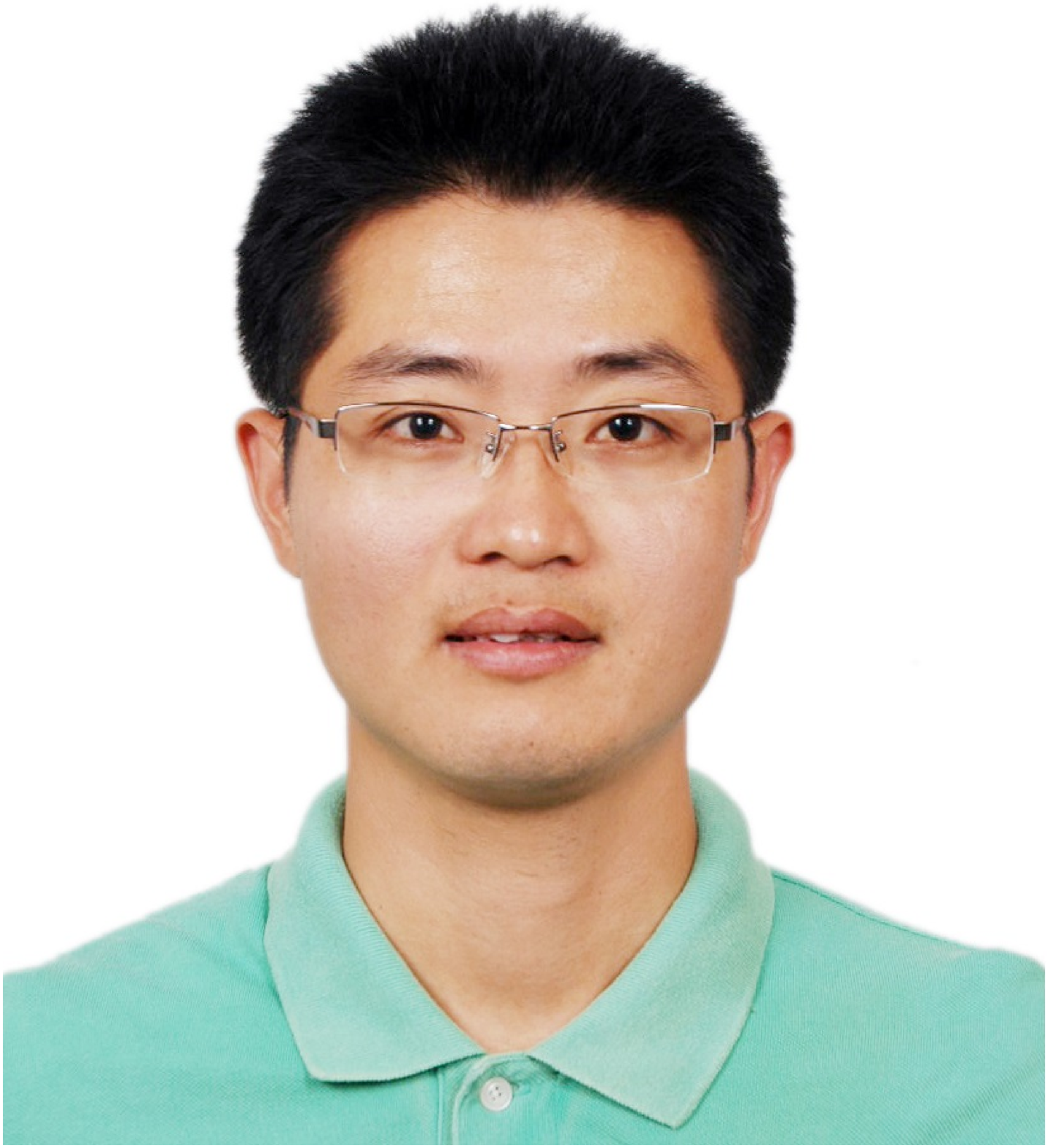}}]{Yingyu Zhang}
received the B.Eng. degree in computer science and technology
from Changsha University of Science and Technology, Changsha, China, in 2002,
and the M.Eng. and Ph.D. degrees in computer science from Huazhong University of Science and Technology, Wuhan, China,
in 2007, and 2011, respectively.
He is now a lecturer with the School of Computer Science, Liaocheng University, Liaocheng, China.
His research interests includes quantum optimization, evolutionary multi-objective optimization, machine learning, and cloud computing.
\end{IEEEbiography}

\begin{IEEEbiography}[{\includegraphics[width=1in,height=1.25in,clip]{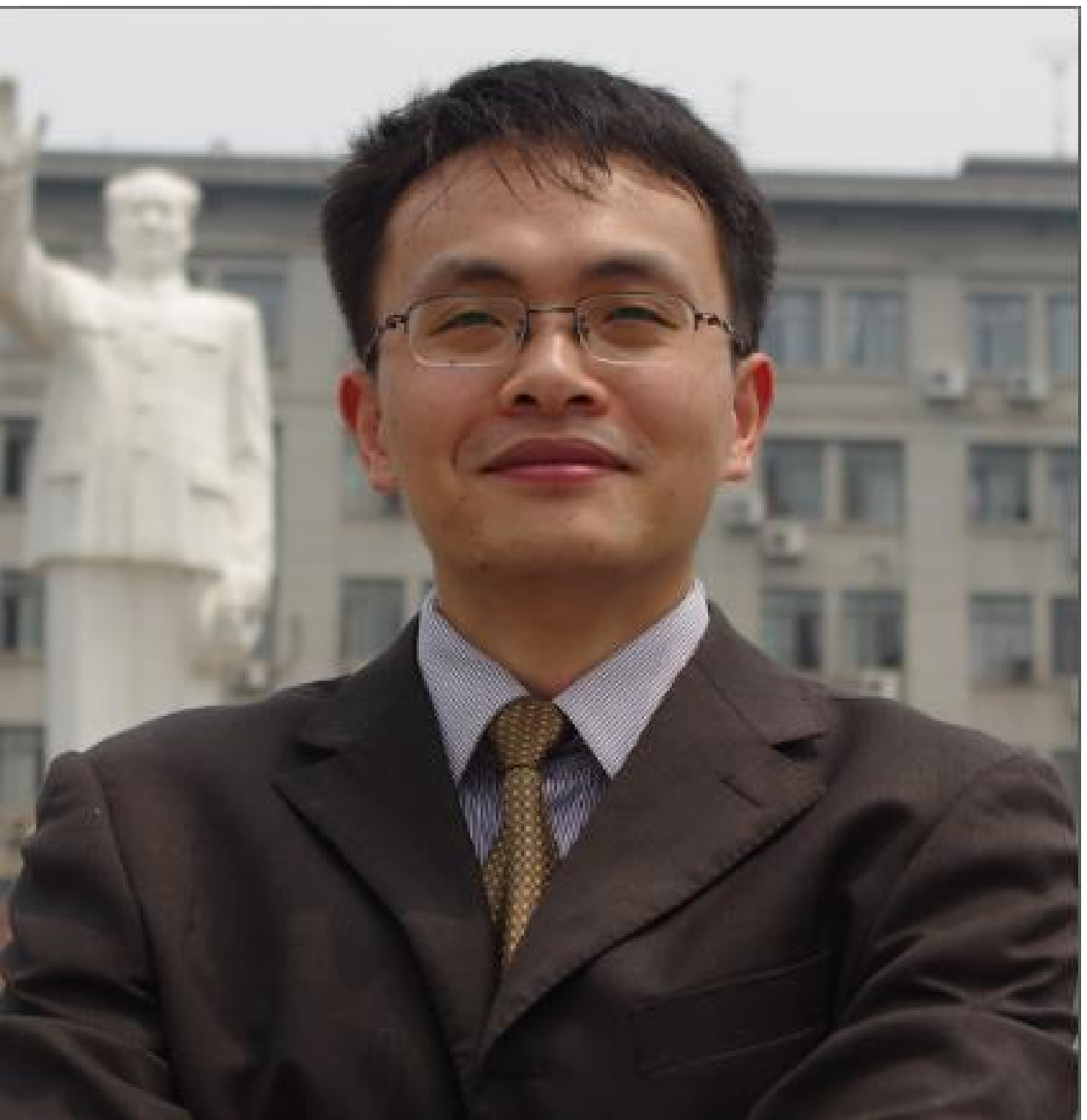}}]{Bing Zeng}
 received the B.Econ. degree in economics from Huazhong Agricultural University, Wuhan, China, in 2004, and received the B.Eng. degree in computer science and technology, the M.Eng. and  Ph.D. degrees in information security from Huazhong University of Science and Technology, Wuhan, China, in 2004, 2007 and 2012, respectively. He is currently an Assistant Professor at South China University of Technology, Guangzhou, China. His research interests are in cryptography and network security, with a focus on secure multiparty computation.
\end{IEEEbiography}

\begin{IEEEbiography}[{\includegraphics[width=1in,height=1.25in,clip]{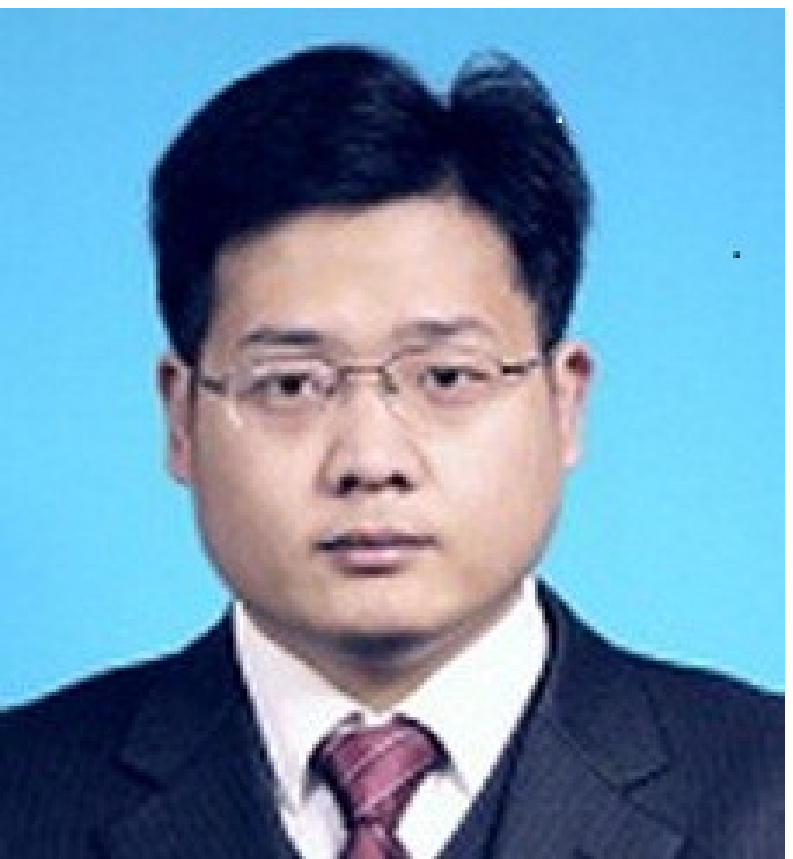}}]{Yuanzhen Li}
 received his PhD degree from Beijing University of Posts and Telecommunications in 2010. He is now an associate professor in the Department of Computer Science and Technology, Liaocheng University, China. His  research interests include wireless communications, evolutionary computation and Multi-objective optimization.
\end{IEEEbiography}

\begin{IEEEbiography}[{\includegraphics[width=1in,height=1.25in,clip]{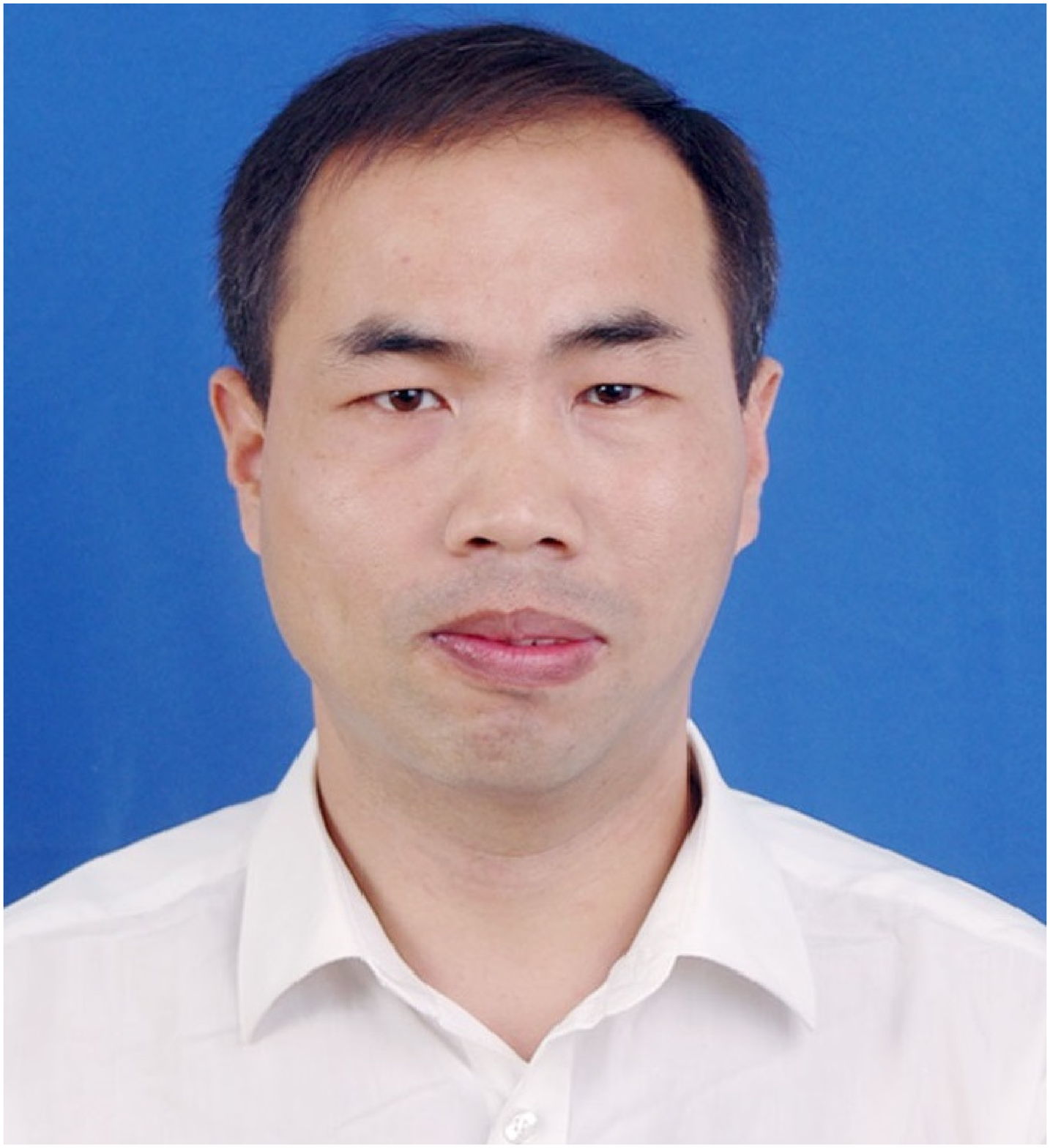}}]{Junqing Li}
received the B.Sc. and Ph.D. degrees from Shandong Economic University, Northeastern University in 2004 and 2016, respectively. Since 2004, he has been with the School of Computer Science, Liaocheng University, Liaocheng, China, where he became an Associate Professor in 2008. He also works with the School of information and Engineering, Shandong Normal University, Jinan, Shandong, China, where he became the doctoral supervisor. His current research interests include intelligent optimization and scheduling.
\end{IEEEbiography}







\end{document}